\documentclass[]{fairmeta}
\usepackage{hyperref}
\usepackage{url}
\usepackage{amsthm}
\usepackage{hyperref}

\usepackage{fontawesome5}
\usepackage[smartEllipses]{markdown}

\usepackage[most]{tcolorbox}
\usepackage{wrapfig}
\usepackage{multirow}
\usepackage{graphicx,xcolor,float}
\usepackage{subfigure}
\usepackage{threeparttable}
\usepackage[ruled,vlined]{algorithm2e}
\usepackage{colortbl}
\usepackage{color}
\usepackage{multirow}
\usepackage{tabularx}
\usepackage{float}
\usepackage{graphicx}
\usepackage{booktabs}
\usepackage{arydshln}
\usepackage{enumitem}
\usepackage{wrapfig}
\usepackage{caption}
\usepackage{graphicx}
\usepackage{makecell}
\usepackage{float} 
\usepackage{mathtools}
\usepackage{hhline}
\usepackage{boldline}
\usepackage{bbding}
\usepackage{makecell}
\usepackage{tabularx}
\usepackage{amssymb,mathrsfs,amsmath}
\usepackage{pifont}
\usepackage{xcolor}
\usepackage{bbm}

\newcommand{\ourmethod}{{\fontfamily{lmtt}\selectfont \textbf{DVD}}\xspace}

\newcommand{\ourmtab}{\fontfamily{lmtt}\selectfont \textbf{DVD}}

\usepackage{fontawesome5}

\usepackage{listings}

\newcommand{\obsbox}[1]{%
    \begin{tcolorbox}[colframe=black!70, colback=yellow!5, boxrule=1pt, arc=2mm,   top=2pt, bottom=3pt, left=3pt, right=3pt,
  boxsep=1pt]
        #1
    \end{tcolorbox}
}

\definecolor{bittersweet}{rgb}{1.0, 0.44, 0.37}
\definecolor{mygreen}{rgb}{0.29, 0.7, 0.48}
\usepackage{pifont}
\definecolor{demphcolor}{RGB}{144,144,144}

\definecolor{mygray}{gray}{0.4}
\definecolor{autopurple}{HTML}{7030A0}
\definecolor{dyna_yellow}{HTML}{BF9000}
\definecolor{adaptive_blue}{HTML}{0070C0}
\definecolor{darksalmon}{rgb}{0.91, 0.59, 0.48}
\definecolor{emerald}{rgb}{0.31, 0.78, 0.47}
\definecolor{green(pigment)}{rgb}{0.0, 0.65, 0.31}
\definecolor{amaranth}{rgb}{0.9, 0.17, 0.31}
\definecolor{iris}{rgb}{0.35, 0.31, 0.81}
\definecolor{uu}{rgb}{0.95, 0.51, 0.51}
\definecolor{spirodiscoball}{rgb}{0.06, 0.75, 0.99}
\usepackage{svg}

\definecolor{mygrey}{gray}{0.4}

\definecolor{QuestionColor}{rgb}{0.7, 0.1, 0.1} 
\definecolor{AnswerColor}{rgb}{0.1, 0.5, 0.1} 
\definecolor{ReasoningColor}{rgb}{0.1, 0.1, 0.7} 

\usepackage[table,xcdraw,usenames,dvipsnames]{xcolor}
\usepackage{cleveref}
\SetKwInOut{Input}{Input}\SetKwInOut{Output}{Output}
\usepackage{twemojis}

\hypersetup{
    colorlinks=true,
    linkcolor=red,
    citecolor=cyan,
    filecolor=magenta,      
    urlcolor=magenta,
    }

\title{DVD: Deterministic Video Depth Estimation with Generative Priors}

\author[1\dagger]{Hongfei Zhang}
\author[1,2\dagger]{Harold Haodong Chen}
\author[1\dagger]{Chenfei Liao}
\author[1\dagger]{Jing He}
\author[1]{Zixin Zhang}
\author[3]{Haodong Li}
\author[4]{~~Yihao Liang}
\author[1]{Kanghao Chen}
\author[5]{Bin Ren}
\author[1]{Xu Zheng}
\author[1]{Shuai Yang}
\author[6]{Kun Zhou}
\author[7]{Yinchuan Li}
\author[8]{Nicu Sebe}
\author[1,2\ddagger]{Ying-Cong Chen}
\affiliation[1]{HKUST(GZ)}
\affiliation[2]{HKUST}
\affiliation[3]{UCSD}
\affiliation[4]{Princeton University}
\affiliation[5]{MBZUAI}
\affiliation[6]{SZU}
\affiliation[7]{Knowin}
\affiliation[8]{UniTrento}
\contribution[\dagger]{Equal Contribution}
\contribution[\ddagger]{Corresponding Author}

\abstract{Existing video depth estimation faces a fundamental trade-off: \textit{generative models} suffer from stochastic geometric hallucinations and scale drift, while \textit{discriminative models} demand massive labeled datasets to resolve semantic ambiguities. To break this impasse, we present \ourmethod, the \textit{first} framework to deterministically adapt pre-trained video diffusion models into single-pass depth regressors. Specifically, \ourmethod features three core designs: (\textbf{\textit{i}}) repurposing the diffusion \textbf{timestep as a structural anchor} to balance global stability with high-frequency details; (\textbf{\textit{ii}}) \textbf{latent manifold rectification (LMR)} to mitigate regression-induced over-smoothing, enforcing differential constraints to restore sharp boundaries and coherent motion; and (\textbf{\textit{iii}}) \textbf{global affine coherence}, an inherent property bounding inter-window divergence, which enables seamless long-video inference without requiring complex temporal alignment. Extensive experiments demonstrate that \ourmethod achieves \textit{state-of-the-art} zero-shot performance across benchmarks. Furthermore, \ourmethod successfully unlocks the profound geometric priors implicit in video foundation models using $163\times$ \textit{less} task-specific data than leading baselines. Notably, we fully release our pipeline, providing the \textit{whole training suite} for SOTA video depth estimation to benefit the open-source community.
 }

\date{March 13, 2026}
\metadata[{\faGlobe\ Project}]{\url{https://dvd-project.github.io/}}
\metadata[{\faGithub\ Github}]{\url{https://github.com/EnVision-Research/DVD}}

\begin{document}

\maketitle

\section{Introduction}
\label{sec:intro}

Depth estimation serves as a fundamental building block for 3D scene understanding, underpinning applications \citep{chen2025guided, charatan2024pixelsplat,xu2025depthsplat, zhang2023controlvideo, o2024open} from autonomous driving to robotic manipulation. While image-based depth estimation has matured significantly~\citep{bochkovskii2024depthpro,yang2024depthv2,yang2024depthv1,piccinelli2024unidepth,yin2023metric3d,fu2024geowizard}, elevating this capability to the video domain remains a formidable challenge. The transition from static images to dynamic video is non-trivial; it demands not only precise geometric reasoning per frame but also rigorous temporal consistency. In real-world scenarios characterized by camera motion and dynamic objects, maintaining this consistency without sacrificing high-frequency geometric details is a persistent bottleneck.

\begin{figure*}[!t]
\centering
\includegraphics[width=\linewidth]{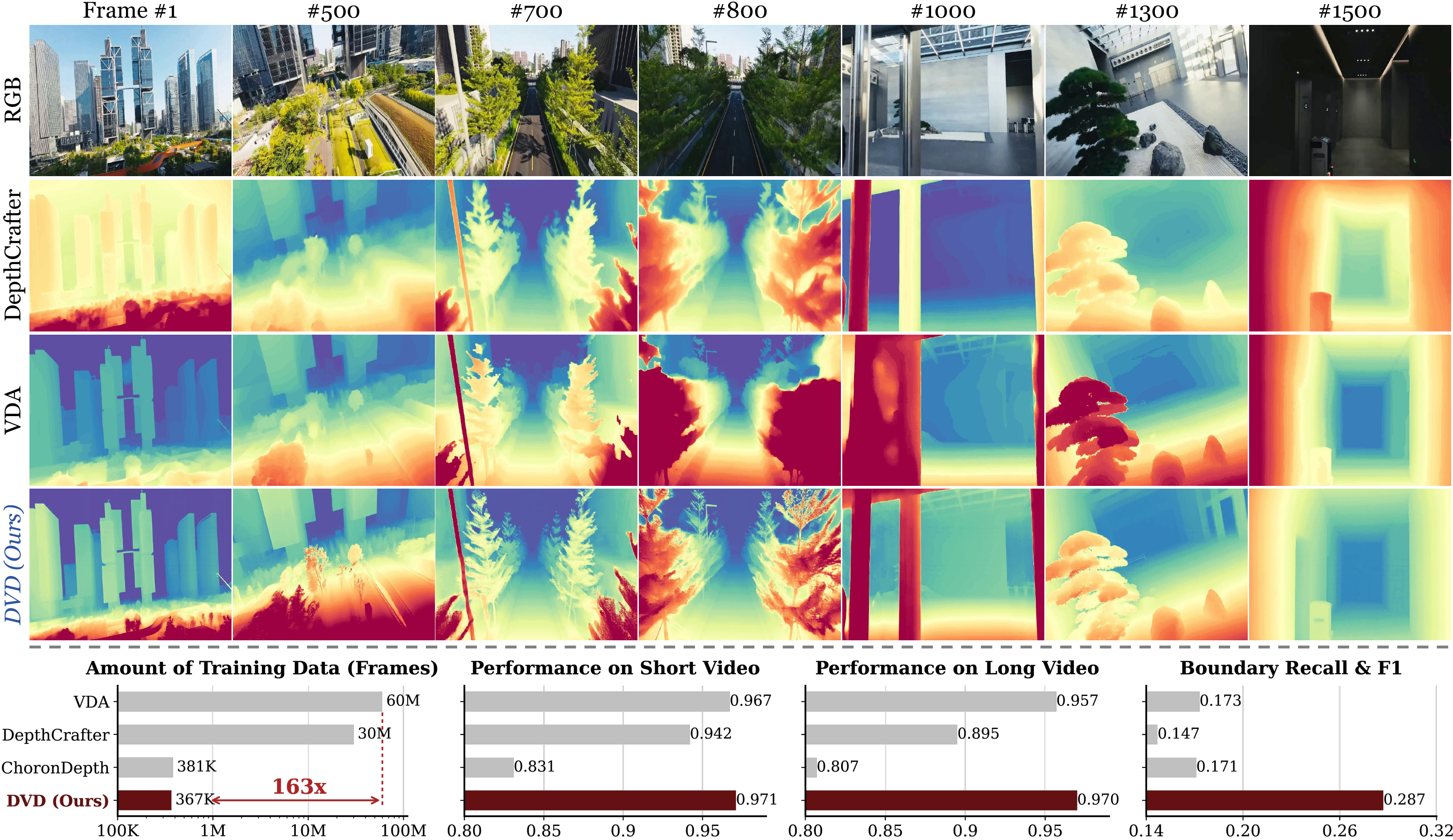}
\vspace{-2em}
\caption{
(\textbf{\textit{Top}}) Comparisons on a $1500$-frame in-the-wild video highlight a fundamental paradigm trade-off: representative generative models (\textit{e.g.}, DepthCrafter~\citep{hu2025depthcrafter}) suffer from \emph{geometric hallucination}, while leading discriminative baselines (\textit{e.g.}, VDA~\citep{chen2025video}) face \emph{semantic ambiguity}. \ourmethod resolves this dilemma, delivering consistent, high-fidelity geometry.
(\textbf{\textit{Bottom}}) \ourmethod achieves superior performance on both short and long videos (averaged on KITTI \citep{Geiger2012CVPR}, ScanNet \citep{dai2017scannet}, and Bonn \citep{palazzolo2019bonn}), while successfully unlocking the rich priors implicit in video foundation models using remarkably minimal task-specific data, \textit{e.g.}, less than $1\%$ of VDA's training set.
}

\label{fig:teaser}
\vspace{-0.8em}
\end{figure*}

Recent advances in video depth estimation have predominantly followed two paradigms, each constrained by inherent limitations that hinder their broader applicability, as shown in Figure~\ref{fig:teaser} (\textit{Top}). \textbf{(I) Diffusion-based generative models} \citep{hu2025depthcrafter,shao2025learning,yang2024depth} (\textit{e.g.}, DepthCrafter) leverage pre-trained video foundation models to capture rich spatio-temporal priors, enabling impressive zero-shot generalization. However, their reliance on stochastic sampling introduces temporal uncertainties which limit their stability and reliability in real-world applications. Moreover, the generative nature of these models tends to prioritize visual plausibility over geometric accuracy, leading to \textbf{geometric hallucination}, a failure to maintain precise and globally consistent geometry over time. \textbf{(II) Discriminative ViT-based models} \citep{yang2024depthv2,chen2025video} (\textit{e.g.}, Video Depth Anything, VDA), on the other hand, provide high inference efficiency and deterministic outputs. Yet, learning geometry strictly from dense annotations, they frequently suffer from \textbf{semantic ambiguity}, misinterpreting motion blur or textureless regions as structural boundaries. To overcome this ambiguity, discriminative paradigms heavily rely on massive-scale and diversified downstream annotations~\citep{chen2025video,yang2024depthv1,yang2024depthv2,birkl2023midas}. This extreme data dependency not only raises significant barriers to scalability and reproducibility but also restricts their adaptability in broader, data-scarce scenarios.
These aforementioned challenges lead to our pivotal research question:

\vspace{-0.4em}
\obsbox{
\includegraphics[height=0.9em]{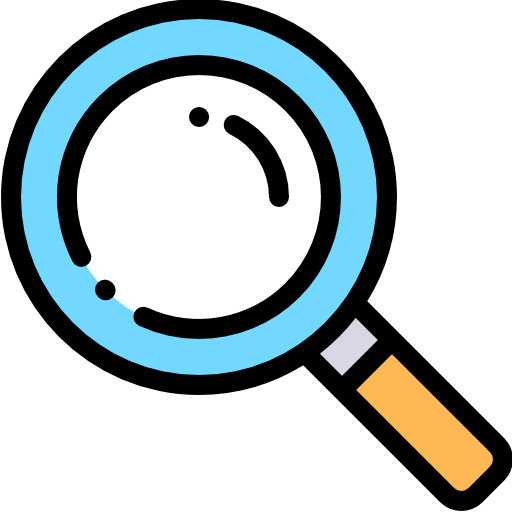}
\textit{Can we design a video depth estimation framework that effectively balances the structural stability of discriminative models and the rich spatio-temporal priors of generative approaches, while remaining efficient and scalable?}
}
\vspace{-0.1em}

In response, we present \ourmethod, a novel framework that achieves deterministic video depth estimation with generative priors.
Departing from the conventional stochastic generative paradigm, \ourmethod pioneers the deterministic adaptation of pre-trained video diffusion models, learning a direct mapping from RGB latents to depth latents.
This paradigm shift introduces a new design point: leveraging the backbone's rich semantic priors to resolve motion-induced ambiguity while enforcing a regression objective that predicts geometrically consistent depth, effectively mitigating generative hallucination. However, extending deterministic adaptation from static images \citep{lee2024exploiting,he2025lotus} to dynamic videos, presents unique challenges: a naive regression is not merely prone to \textit{blurring}, but suffers from \textit{structural instability} and \textit{scalability} issues \citep{hu2025depthcrafter,shao2025learning}.

To address these bottlenecks, \ourmethod introduces a video deterministic adaptation paradigm built upon three key mechanisms:
\ding{182} \textbf{Timestep as a Structural Anchor:} We repurpose the diffusion timestep $t$ from a noise-level index into a structural anchor, effectively balancing low-frequency geometric stability with high-frequency spatial details. \ding{183} \textbf{Latent Manifold Rectification (LMR):} To combat the fundamental spatio-temporal "mean collapse" in deterministic regression, we introduce a parameter-free supervision that aligns latent differentials, successfully restoring sharp boundaries and coherent motion. \ding{184} \textbf{Global Affine Coherence:} We uncover that our deterministic backbone inherently bounds inter-window divergence. This property enables a seamless, affine-alignment sliding-window inference strategy for long-duration videos, bypassing complex latent stitching~\citep{hu2025depthcrafter, shao2025learning}. Ultimately, \ourmethod resolves the ambiguity-hallucination dilemma by successfully repurposing video generation models into deterministic regressors. This paradigm shift achieves \textit{state-of-the-art} zero-shot video depth estimation. Notably, \ourmethod effectively unlocks the rich geometric priors embedded in foundation models using remarkably minimal task-specific data. This establishes a highly efficient and scalable adaptation route for future 3D perception. In brief, our contributions are summarized as follows:

\begin{itemize}[leftmargin=1.6em]
    \item[\ding{111}] \textbf{Bottleneck Identification.} 
    Through analysis of existing video depth estimation paradigms, we identify key bottlenecks: geometric hallucination in generative models and semantic ambiguity in discriminative models, which hinder scalability and practical deployment.
    \item[\ding{111}] \textbf{Our Solution.} We present \ourmethod, which pioneers the deterministic adaptation of pre-trained video diffusion models into single-pass regressors. \ourmethod leverages three key insights to address the identified bottlenecks: (\textbf{\textit{i}}) timestep as a structural anchor, which balances the trade-off between geometric stability and detail precision; (\textbf{\textit{ii}}) latent manifold rectification, which ensures spatial and temporal consistency; and (\textbf{\textit{iii}}) global affine coherence, which enables robust long-video inference.
    \item[\ding{111}] \textbf{Empirical Validation.} Extensive experiments across four real-world benchmarks demonstrate that \ourmethod achieves \ding{182} \textbf{superior performance:} achieving \textit{state-of-the-art} zero-shot geometric fidelity and temporal coherence; \ding{183} \textbf{compelling efficiency:} successfully unlocking pre-trained world priors with remarkably minimal downstream data (\textit{e.g.}, $<1\%$ of leading baselines) while maintaining comparable inference speed; and \ding{184} \textbf{robust scalability:} enabling seamless, robust inference on long videos and effortlessly generalizing to unconstrained open-world domains.
\end{itemize}

\section{Related Work}

\textbf{Monocular Depth Estimation.}\quad Modern approaches have evolved from handcrafted features to data-driven deep learning~\citep{bhat2021adabins,eigen2014depth}, broadly categorizing into two dominant paradigms: \textbf{(I)~Discriminative Regression:} This paradigm leverages ViTs and large-scale supervision to learn direct depth mappings~\citep{chou2025flashdepth,hu2024metric3d,wang2025moge,birkl2023midas,sobko2026stabledpt,piccinelli2025video}. Foundation models like Depth Anything V1/V2 \citep{yang2024depthv1,yang2024depthv2,pham2025sharpdepth} demonstrate robust zero-shot generalization by scaling up unlabeled pre-training. To recover metric scale, approaches such as Metric3D \citep{yin2023metric3d}, UniDepth \citep{piccinelli2024unidepth}, and Depth Pro \citep{bochkovskii2024depthpro} focus on resolving focal length ambiguities and preserving high-frequency details. In the video domain, methods like Video Depth Anything \citep{chen2025video} extend these backbones with temporal modules or flow-based refinement. While efficient and deterministic, discriminative methods typically lack the generative priors necessary to resolve semantic ambiguities in textureless or motion-blurred regions.
\textbf{(II) Generative Diffusion:} To incorporate rich geometric priors, recent works repurpose pre-trained diffusion models for depth estimation~\citep{song2025depthmaster,kong2025worldwarp,li2025stereodiff,ranftl2021vision,bhat2023zoedepth,yang2024depth}. Image-based methods, such as Marigold \citep{ke2025marigold} and Lotus \citep{he2024lotus,he2025lotus}, fine-tune latent diffusion models to achieve superior structural detail compared to discriminative baselines. Video-specific approaches, including ChronoDepth \citep{shao2025learning}, DepthCrafter \citep{hu2025depthcrafter}, and RollingDepth \citep{ke2025video}, further adapt these priors to model temporal dynamics.
However, their reliance on stochastic multi-step sampling inherently introduces high latency and geometric hallucinations, a bottleneck our \ourmethod resolves via deterministic adaptation.

\textbf{Video Diffusion Models.}\quad The field has witnessed a paradigm shift from adapting 2D U-Nets \citep{ronneberger2015u} to scalable diffusion transformers (DiT) \citep{peebles2023scalable}.
Early pioneering works~\citep{blattmann2023stable, blattmann2023align, ho2022video, guo2023animatediff, wang2023modelscope} primarily extended pre-trained image architectures by inserting temporal attention or 3D convolutions.
Recently, a paradigm shift has occurred driven by spacetime patchified sequence modeling \citep{brooks2024video} and continuous flow matching \citep{lipman2022flow}.
By scaling DiT with advanced 3D VAEs, modern foundation models \citep{genmo2024mochi,seedance2025seedance,gao2025seedance, chen2025skyreels}, such as CogVideoX~\citep{hong2022cogvideo, chen2025hierarchical, chenevent}, HunyuanVideo~\citep{kong2024hunyuanvideo}, and Wan~\citep{wan2025wan, zhang2025dualcamctrl}, have demonstrated unprecedented capabilities in simulating physical dynamics and maintaining strict 3D consistency \citep{huang2024vbench, chen2025tivibench}.
These foundation models effectively function as world simulators, encoding rich geometric and dynamic priors that \ourmethod repurposes for deterministic depth regression.

\noindent\textbf{More Video Depth Methods.}\quad
Beyond the generative and discriminative paradigms for relative video depth discussed above, recent advancements have diversified video depth estimation into several specialized tracks. One prominent direction optimizes for \textbf{(I) real-time streaming efficiency}, with methods like FlashDepth~\citep{chou2025flashdepth} and VeloDepth~\citep{piccinelli2025video} employing lightweight architectures for latency-critical applications. Other parallel tracks like \textbf{(II) metric geometry recovery}, where GeometryCrafter~\citep{xu2025geometrycrafter} alters the target representation to unbounded point maps to facilitate downstream 3D/4D reconstruction. While these works make significant strides in their respective settings, their primary objectives diverge fundamentally from our problem formulation, where \ourmethod explores a more general direction for video depth. Consequently, these works also serve as valuable complementary approaches to the field, rather than direct baselines for our core setting.

\section{Preliminary}
\vspace{-0.2em}

\subsection{Problem Formulation}

We formalize video depth estimation as a mapping from an input RGB sequence $x \in \mathbb{R}^{F \times 3 \times H \times W}$ to its corresponding depth sequence $d \in \mathbb{R}^{F \times H \times W}$, where $F$ denotes the frame count. 
To exploit the rich semantic priors of large-scale pre-trained models, we operate within a compressed latent manifold.
Specifically, a frozen variational autoencoder (VAE) encoder $\mathcal{E}(\cdot)$ projects both RGB and depth into a unified latent space:
\setlength\abovedisplayskip{3pt}
\setlength\belowdisplayskip{3pt}
\begin{equation}
    z_x = \mathcal{E}(x) \in \mathbb{R}^{f \times C \times h \times w}, \quad 
    z_d = \mathcal{E}(d) \in \mathbb{R}^{f \times C \times h \times w},
\end{equation}
where $c, f, h, w$ represent the latent channels and downsampled dimensions, respectively.
Our objective is to learn a deterministic mapping $\Phi: z_x \mapsto z_d$ that recovers the geometric structure directly in the latent space. The final depth $\hat{d}$ is reconstructed via the frozen VAE decoder $\hat{d} = \mathcal{D}(\hat{z}_d)$.

\begin{figure}[!t]
    \centering
    \includegraphics[width=\linewidth]{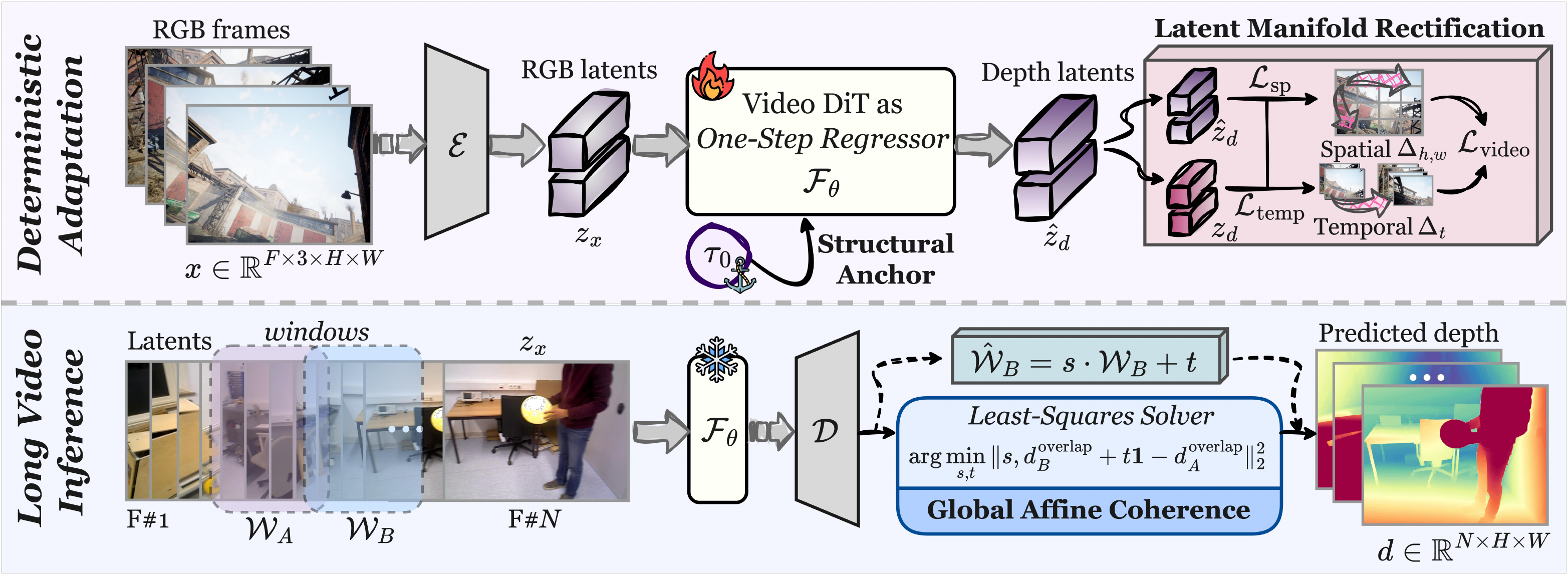}
    \vspace{-2em}
    \caption{
    \textbf{Overview of  \ourmethod.} (\textbf{\textit{Top}}) A video DiT ($\mathcal{F}_\theta$) performs single-pass depth regression, modulated by a structural anchor ($\tau_0$). Latent manifold rectification (LMR) mitigates mean collapse via differential constraints. (\textbf{\textit{Bottom}}) For long video depth estimation, overlapping windows ($\mathcal{W}_A, \mathcal{W}_B$) are seamlessly aligned using a closed-form least-squares solver, leveraging the model's global affine coherence.
    }
    \vspace{-0.8em}
    \label{fig:framework}
\end{figure}

\subsection{Diffusion as Deterministic Regressor}

\noindent\textbf{Role of $t$ in Rectified Flow.}\quad
In traditional rectified flow (RF)~\citep{liu2022flow,lipman2022flow}, the time variable $t \in [0,1]$ explicitly parameterizes a noise interpolation trajectory
between data distribution $z_0 \sim p_{\text{data}}$ and Gaussian noise $z_1 \sim \mathcal{N}(0, I)$.
RF defines a linear interpolation trajectory $z_t = (1 - t) z_0 + t z_1$, where the scalar timestep $t \in [0, 1]$ explicitly parameterizes the corruption level.
The network $v_\theta$ is trained to predict the velocity field of this flow by minimizing:
\begin{equation}
\mathcal{L}_{\text{RF}} = \mathbb{E} [ \| v_\theta(z_t, t) - (z_1 - z_0) \|^2 ].
\label{eq:L2}
\end{equation}
During standard generative inference, one produces samples by solving the ordinary differential equation (ODE) $d z_t / dt = v_\theta(z_t, t)$ via numerical integration from $t=1$ to $t=0$.

\noindent\textbf{Deterministic Adaptation.}\quad 
Recent works in the image domain \citep{he2024lotus,he2025lotus} repurpose diffusion backbones as \textit{one-step deterministic regressors}. Instead of iterative ODE integration over a noise trajectory, the network $\mathcal{F}_\theta$ performs a direct functional mapping. Formally, given the RGB latent $z_x$ and a timestep condition $t$, depth is deterministically predicted in a single forward pass:
\begin{equation}
\hat{z}_d = \mathcal{F}_\theta(z_x, t).
\label{eq:deterministic_mapping}
\end{equation}
Building upon this static-image formulation, Section~\S\ref{sec:method} details how \ourmethod extends this paradigm to videos, along with uncovering a crucial functional shift for the timestep $t$ to preserve geometric consistency.

\section{Methodology}
\label{sec:method}
\vspace{-0.2em}

\subsection{Overall Framework}
Existing video depth estimation methods are typically polarized: generative diffusion models offer rich spatio-temporal priors but suffer from stochastic geometric hallucinations, while discriminative regressors provide stable outputs but demand massive labeled datasets to resolve semantic ambiguities.
Our \ourmethod is proposed to bridge this gap, a novel framework that unites the generalization power of generative priors with the structural stability of deterministic regression, as shown in Figure~\ref{fig:framework}.
Formally, given an input RGB video $x$, a VAE encoder $\mathcal{E}$ extracts the latent representation $z_x$. This latent sequence is then processed by a pre-trained video diffusion backbone $\mathcal{F}_\theta$. Instead of performing iterative stochastic denoising, \ourmethod executes a single-pass deterministic mapping to predict the depth latent $\hat{z}_d$, modulated by a conditioning timestep $\tau$:
\begin{equation}
    \hat{z}_d = \mathcal{F}_\theta(z_x, \tau(t)).
\end{equation}
To achieve high-fidelity depth estimation, \ourmethod introduces three core designs tailored to the latent dynamics of video diffusion backbones. First, we repurpose the diffusion timestep $\tau$ as a \textit{structural anchor} (Section~\S\ref{sec:timeanchor}) to govern the backbone's geometric operating regime, balancing low-frequency stability with high-frequency details. Then, we introduce \textit{latent manifold rectification} (Section~\S\ref{sec:rectification}), a parameter-free supervision mechanism that enforces differential consistency to mitigate regression-induced mean collapse and sharpen spatio-temporal boundaries. Finally, we present \textit{global affine coherence} (Section~\S\ref{sec:scale}), an inherent property of our deterministic backbone that strictly bounds inter-window divergence, enabling seamless, affine-alignment inference for long-duration videos. We next detail the empirical observations and technical formulations that motivate these designs in the following sections.

\subsection{Timestep as Structural Anchor}
\label{sec:timeanchor}

\begin{figure*}[!t]
\centering
\includegraphics[width=\linewidth]{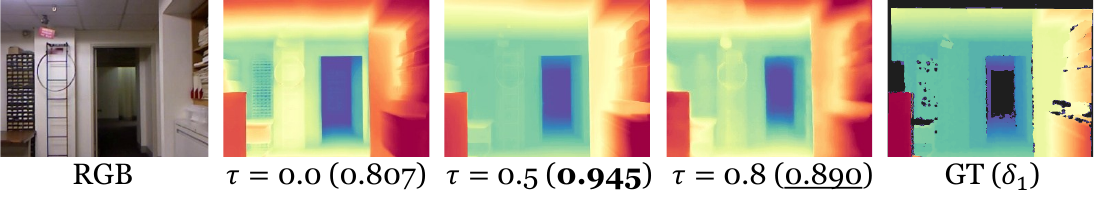}
\vspace{-2.7em}
\caption{
\textbf{Timestep as a structural anchor.} Visualizations on NYU~\citep{SilbermanECCV12nyuv2} demonstrate a fidelity-stability trade-off. Low ($\tau=0.0$) recovers sharp boundaries but lacks global consistency, whereas high ($\tau=0.8$) causes detail loss (\textit{e.g.}, blur). An optimal anchor ($\tau=0.5$) balances these regimes, achieving a trade-off between detail recovery and metric accuracy. More detailed quantitative analyses are shown in Figure~\ref{fig:ablation_timestep}.
}
\label{fig:vis_tau}
\vspace{-0.8em}
\end{figure*}

In single-image deterministic adaptation (\textit{e.g.}, Lotus~\citep{he2024lotus, he2025lotus}), the diffusion timestep is typically fixed at the terminal state ($t=1$) or absorbed entirely. However, we empirically observe that applying this to video backbones causes severe geometric over-smoothing (Figure~\ref{fig:vis_tau}). 
We attribute this to the spectral bias inherent in the pre-trained diffusion priors~\citep{kingma2021variational,choi2022perception,hang2025improved,hang2023efficient,ho2022video}. During generative pre-training, the timestep $t$ parameterizes the signal-to-noise ratio (SNR): a higher $t$ (early time, low SNR) forces the network to estimate low-frequency global structures, while a lower $t$ (late time, high SNR) trains the network to resolve high-frequency local details. Therefore, in our deterministic adaptation, the timestep transcends its traditional role as a noise indicator. By replacing the dynamic timestep $t$ with a conditioning state $\tau_0$, we instantiate a persistent \textbf{structural anchor} that explicitly modulates the network's geometric operating regime.

\noindent\textbf{Frequency-Parameterized Conditioning.}\quad
To better understand this mechanism, we analyze how $t$ enters the network. In \ourmethod, $t$ acts as a frequency-parameterized condition via a fixed sinusoidal basis~\citep{kim2024disappearancetimestepembeddingmodern,wan2025wan}:
\begin{equation}
    \mathbf{e}_{\text{sin}}(t) = \big[ \cos(\omega_1 t), \ldots, \cos(\omega_{d/2} t), \sin(\omega_1 t), \ldots, \sin(\omega_{d/2} t) \big],
    \label{eq:cos_sin_emb}
\end{equation}
where $d$ is the embedding dimension and $\{\omega_i\}$ are predefined angular frequencies. Rather than sampling $t \sim \mathcal{U}(0,1)$, \ourmethod anchors the model to a single optimal state $\tau_0$. This instantiates a persistent structural code that calibrates the backbone's conditioning pathway. The deterministic mapping is thus formulated as:
\begin{equation}
    \hat{z}_d = \mathcal{F}_\theta\left(z_x; \, \mathbf{e}_\phi(\tau_0)\right),
\end{equation}
where $\mathbf{e}_\phi(\cdot)$ denotes the MLP projection of the sinusoidal embedding.

\begin{wrapfigure}{r}{0.5\textwidth}
\vspace{-2.2em}
 \centering
 \includegraphics[width=\linewidth]{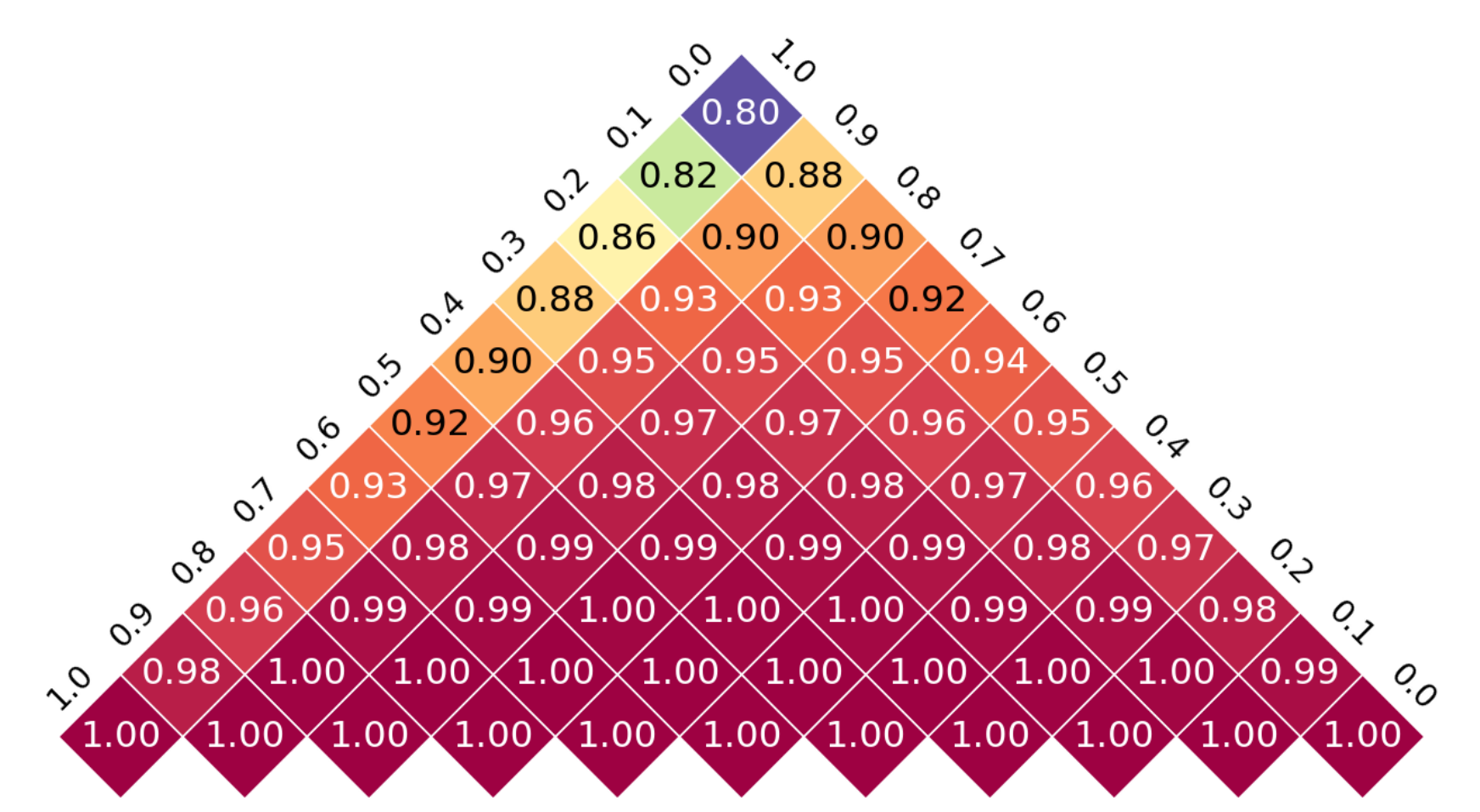}
  \vspace{-2.1em}
  \caption{\textbf{Timestep embedding similarity.}
  Cosine similarity matrix of timestep embeddings ($t \in [0, 1]$, stride $0.1$). While embeddings are broadly consistent, mid-range timesteps exhibit high similarity with a wider range of states.  
  }
  \vspace{-2em}
  \label{fig:freq}
\end{wrapfigure}

\noindent\textbf{Fidelity-Stability Trade-off.}\quad
Our key finding is that the choice of $\tau_0$ induces a strict \textit{fidelity-stability trade-off} that persists even after fine-tuning converges.
As shown in~Figure~\ref{fig:vis_tau}, early timestep~(\textit{e.g.}, $\tau = 0.8$) biases the model toward low-frequency global structures (stable but blurry), while late timestep~(\textit{e.g.}, $\tau = 0.0$) amplifies high-frequency details (sharp but unstable). Among the broadly similar embeddings in Figure~\ref{fig:freq}, anchoring at mid-range timestep~(\textit{e.g.}, $\tau=0.5$) offers a low-variation conditioning region, better balancing global coherence with local sharpness~(see Figure~\ref{fig:ablation_timestep} for more details). We further found that completely ablating this conditioning or re-initializing significantly degrades performance, confirming $\tau$ indexes an irreplaceable pre-trained geometric prior~(Appendix~\S\ref{apx:more_ablation}).

\subsection{Latent Manifold Rectification}
\label{sec:rectification}

\begin{wrapfigure}{r}{0.5\textwidth}
\vspace{-1.7em}
 \centering
 \includegraphics[width=\linewidth]{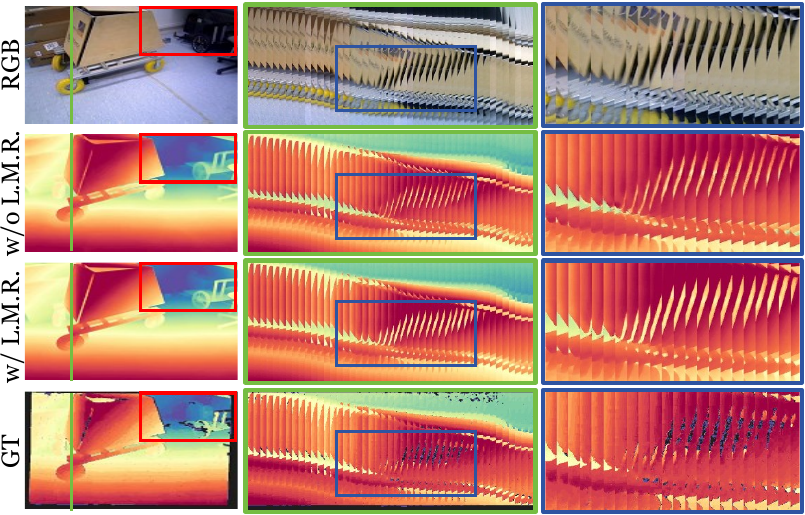}
  \vspace{-2em}
  \caption{\textbf{LMR mitigates mean collapse.}
  Naive regression (\textit{\textbf{2nd Row}}) exhibits mean collapse, losing high-frequency details. In contrast, our LMR (\textit{\textbf{3rd Row}}) enforces differential constraints to rectify the latent manifold, recovering both sharp spatial boundaries and temporal coherence. Quantitative analyses are placed in Figure~\ref{fig:ablation}.
}
  \vspace{-2.2em}
  \label{fig:lmr}
\end{wrapfigure}

While anchoring the timestep at $\tau_0$ establishes an optimal operating regime for the backbone, training a diffusion-based deterministic regressor with point-wise objectives (\textit{e.g.}, $\mathcal{L}_2$) introduces a fundamental limitation, which we term \textbf{mean collapse}. 
Specifically, minimizing a point-wise loss to map RGB latents $z_x$ to depth latents $z_d$ inherently drives the predictor toward the conditional expectation $\mathbb{E}[z_d|z_x]$ \citep{ma2025neural,song2020denoising,song2020score,liu2022flow}. In ambiguous or occluded regions, this regression-to-the-mean forcefully collapses multi-modal geometric hypotheses~\citep{papyan2020prevalence,zhu2021geometric}, washing out high-frequency structural details. Notably,  this degradation is further amplified under the spatio-temporal setting: the suppressed high-frequency differentials propagate and accumulate
temporally, manifesting as progressive boundary erosion and severe motion flickering, as illustrated in Figure~\ref{fig:lmr}.

\noindent\textbf{Differential Manifold Constraints.}\quad
To counteract this regression-induced mean collapse without introducing heavy auxiliary modules~\citep{he2025lotus}, we propose \textbf{latent manifold rectification (LMR)}, a lightweight, \emph{parameter-free} supervision strategy that restores the local differential geometry of the prediction in the VAE latent space. LMR enforces first-order consistency between the predicted and target latents by aligning their spatial and temporal differentials. This mechanism explicitly preserves the differential statistics (gradient and flow distributions) that are typically erased by standard regression, successfully restoring sharp boundaries and coherent motion dynamics~(see Figure~\ref{fig:lmr}).

\begin{itemize}[leftmargin=1.6em]
    \item[\ding{111}] \textbf{Spatial Rectification (Latent Gradient).}
    To preserve sharp geometric discontinuities encoded within the latent space, \ourmethod aligns the spatial gradient fields using finite differences $\nabla_h, \nabla_w$:
    \begin{equation}
        \mathcal{L}_{\text{sp}} = \frac{1}{F \cdot \Omega} \sum_{f=1}^{F} \sum_{\partial \in \{\nabla_h, \nabla_w\}} \| \partial \hat{z}_d^f - \partial z_d^f \|_1,
    \end{equation}
    where $\Omega$ is the spatial resolution. This explicitly penalizes low-frequency latent collapse, enforcing the recovery of fine-grained structural boundaries.

    \item[\ding{111}] \textbf{Temporal Rectification (Latent Flow).}
    Temporal artifacts correspond to mismatched dynamics in the latent depth manifold. \ourmethod therefore synchronizes the predicted temporal flow with ground-truth dynamics:
    \begin{equation}
        \mathcal{L}_{\text{temp}} = \frac{1}{(F-1) \cdot \Omega} \sum_{f=2}^{F} \| \nabla_t \hat{z}_d^f - \nabla_t z_d^f \|_1,
        \label{eq:temp_loss}
    \end{equation}
    where $\nabla_t z^f = z^f - z^{f-1}$. By constraining inter-frame differentials, $\mathcal{L}_{\text{temp}}$ suppresses stochastic mode switching and preserves consistent motion flow.
\end{itemize}

\noindent
The overall objective integrates differential rectification with global consistency:
\begin{equation}
    \mathcal{L}_{\text{video}} = \|\hat{z}_d - z_d\|_2 + \lambda_\text{sp} \mathcal{L}_{\text{sp}} + \lambda_\text{temp} \mathcal{L}_{\text{temp}}.
\end{equation}
Here, $\mathcal{L}_2$ anchors the global geometry, while LMR terms act as a vital safeguard, preserving latent high-frequency structures and temporal dynamics against the smoothing effects of deterministic regression, as shown in Figure~\ref{fig:ablation}. Additional ablations supporting the role of LMR for deterministic regression are provided in Appendix~\S\ref{apx:more_ablation}.

\subsection{Global Affine Coherence}
\label{sec:scale}

\begin{wrapfigure}{r}{0.37\textwidth}
\vspace{-2.2em}
 \centering
 \includegraphics[width=\linewidth]{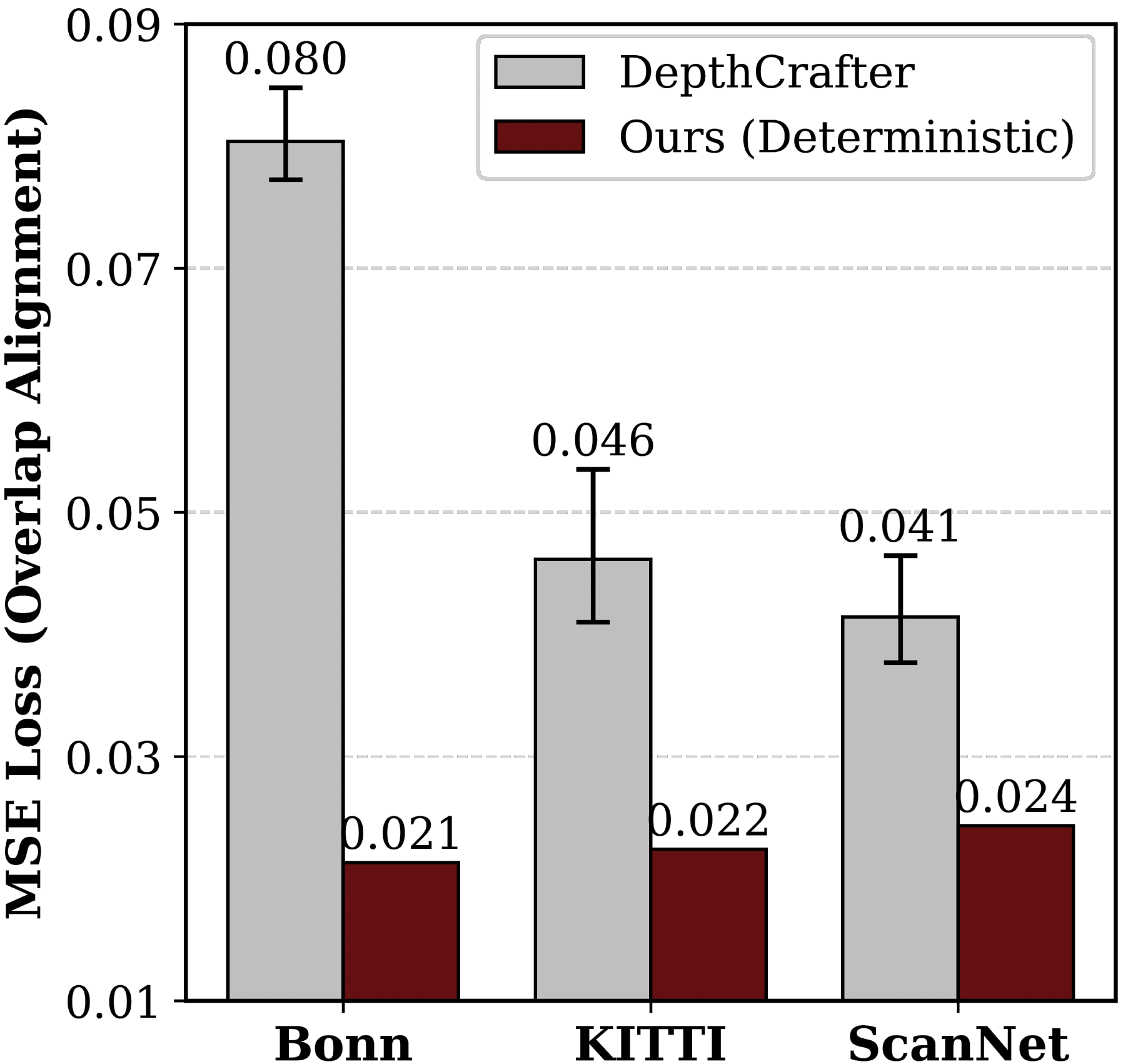}
  \vspace{-2em}
  \caption{
\textbf{Inter-window overlap consistency} \citep{Geiger2012CVPR,palazzolo2019bonn,dai2017scannet}. Unlike generative baselines with high alignment error and variance, our deterministic regression yields minimal MSE and zero variance, validating our global affine coherence that bounds inter-window discrepancies to linear transformations.
}
  \vspace{-2.2em}
  \label{fig:long_vid_illustration}
\end{wrapfigure}
While LMR ensures high-fidelity spatio-temporal consistency within a given sequence, processing long videos introduces a new challenge: memory constraints necessitate sliding-window inference. In this regime, generative diffusion models inevitably suffer from \textit{stochastic scale drift}~\citep{hu2025depthcrafter,shao2025learning}. Their independent probabilistic sampling across windows causes non-linear geometric deformations and severe flickering, as illustrated in Figure~\ref{fig:teaser}. By contrast, \ourmethod operates as a deterministic regressor ($\mathrm{Var}[\hat{z}_d \mid z_x] = 0$), fundamentally eliminating uncertain output.

\noindent\textbf{Global Affine Coherence.}\quad
Despite this deterministic stability in the latent space, naive windowed inference encounters a secondary bottleneck during pixel decoding. The VAE decoder's context-dependent normalization inevitably induces fluctuations of depth value. Crucially, we empirically uncover a strong \textbf{global affine coherence} within our backbone. 
In practice, VAE decoding predominantly induces global affine variations rather than local spatial distortions, so that inter-window discrepancies can be well-approximated by a linear scale-shift transformation~(Equations~(\ref{eq:least_squares}) and (\ref{eq:scale_shift})).
As validated in Figure~\ref{fig:long_vid_illustration}, this mapping aligns adjacent windows with minimal residual error. Unlike generative models suffering from unalignable stochastic distortions, our divergence is more predictable and mathematically recoverable. We further discuss the boundary conditions of this empirical observation in Appendix~\S\ref{app:failure}.

\noindent\textbf{Long-Video Inference.}\quad
Exploiting this well-bounded, affine-invariant property, we propose a lightweight, parameter-free \textit{affine-alignment} strategy for sliding-window inference. Let $\mathcal{W}_A$ and $\mathcal{W}_B$ denote the decoded depth tensors for the preceding and current windows, respectively. \ourmethod extracts the flattened depth predictions $\mathbf{d}_A^{\mathrm{overlap}}, \mathbf{d}_B^{\mathrm{overlap}} \in \mathbb{R}^{N}$ exclusively from their $N$ overlapping pixels. To align $\mathcal{W}_B$ to the canonical scale of $\mathcal{W}_A$, \ourmethod estimates a global scale $s$ and shift $t$ by minimizing the least-squares objective over the overlap:
\begin{equation}
    \arg\min_{s,t} \| s \mathbf{d}_B^{\mathrm{overlap}} + t\mathbf{1} - \mathbf{d}_A^{\mathrm{overlap}} \|_2^2.
    \label{eq:least_squares}
\end{equation}
This yields a deterministic closed-form solution:
\begin{equation}
    s = \frac{\text{Cov}(\mathbf{d}_A^{\mathrm{overlap}}, \mathbf{d}_B^{\mathrm{overlap}})}{\text{Var}(\mathbf{d}_B^{\mathrm{overlap}})}, \quad t = \mu_A - s \mu_B,
    \label{eq:scale_shift}
\end{equation}
where $\mu_A$ and $\mu_B$ denote the mean values of the overlapping regions. This single affine calibration is then broadcast to the \textit{entire} current window ($\hat{\mathcal{W}}_B = s \cdot \mathcal{W}_B + t$) and smoothly blends the overlapping frames via linear interpolation. This strategy enables seamless, long video inference without requiring complex feature matching, flow estimation, or recurrent temporal modules (\textit{e.g.}, in \citep{hu2025depthcrafter, yang2024depth,shao2025learning}).

\begingroup
\setlength{\tabcolsep}{5.6pt}
\begin{table*}[t!]
\renewcommand{\arraystretch}{1.2}
\centering
\caption{\textbf{Zero-shot video depth estimation results}. The \textbf{best} and the \underline{second best} results are highlighted.}
\vspace{-0.8em}
\footnotesize
\begin{tabular}{lcccccccccccc}
\hlineB{2.5}
\centering
\multirow{2}{*}{\textbf{Method}} & 
\multirow{2}{*}{\textbf{Train Frames}} \hspace{-0.2em} &
\multicolumn{2}{c}{\textbf{KITTI}} & 
\multicolumn{2}{c}{\textbf{ScanNet}} & 
\multicolumn{2}{c}{\textbf{Bonn}} &
\multicolumn{2}{c}{\textbf{Sintel}} \\ 
\cmidrule(lr){3-4}\cmidrule(lr){5-6}\cmidrule(lr){7-8}\cmidrule(lr){9-10}   
& 
& \textbf{AbsRel$\downarrow$} & \textbf{$\delta_1$$\uparrow$} 
& \textbf{AbsRel$\downarrow$} & \textbf{$\delta_1$$\uparrow$}
& \textbf{AbsRel$\downarrow$}  & \textbf{$\delta_1$$\uparrow$}  
& \textbf{AbsRel$\downarrow$}  & \textbf{$\delta_1$$\uparrow$} 
\\
\hlineB{1.5}

DAv2-L~\citep{yang2024depthv2}& -
& 10.9 & 0.913
& 6.4 & 0.967 
& 6.9 & 0.957 
& 50.0 & 0.557
 \\
Marigold v1.1~\citep{ke2025marigold}& -
& 9.5 & 0.936
& 7.6 & 0.940 
& 9.5 & 0.936
& 65.1 & 0.411
 \\
RollingDepth~\citep{ke2024rollingdepth}& -
& 9.8 & 0.912
& 5.8 & 0.964 
& 5.9 & 0.966
& 43.7 & 0.500
 \\
  \hline

ChoronDepth~\citep{shao2025learning}& 381K 
& 15.2 & 0.775
& 17.1 & 0.818 
& 16.8 & 0.901 
& 52.8 & 0.504
 \\

DepthCrafter~\citep{hu2025depthcrafter}& $\sim$ 30M 
& 9.9  & 0.907
& 7.1  & 0.960  
& 5.9  & 0.959
& \textbf{37.1} & \underline{0.664}\\


VDA~\citep{chen2025video}     
& 60M 
& {7.2} & {0.963} 
& \underline{5.8} & \underline{0.968} 
& \textbf{4.7} & \underline{0.970} 
& \underline{39.7} & 0.654
\\

\hline 
\rowcolor{CadetBlue!20} 
\textbf{\ourmtab~(Ours)}   
& 367K & 
\textbf{6.7} & \textbf{0.967} & 
\textbf{5.5} & \textbf{0.974} & 
\textbf{4.7} & \textbf{0.971} &
{44.5} & \textbf{0.667}
\\

 \hlineB{2.5}
\end{tabular}
\label{tab:short_video}
\vspace{-0.8em}
\end{table*}

\begingroup
\setlength{\tabcolsep}{10.4pt}
\begin{table*}[t!]
\renewcommand{\arraystretch}{1.2}
\centering
\caption{\textbf{Zero-shot long video depth estimation results.} Paradigm denotes the backbone type and inference paradigm, where "Diff.", "ViT", "D", and "G" denote Diffusion-based, ViT-based, Discriminative, and Generative, respectively.}
\vspace{-0.8em}
\footnotesize
\begin{tabular}{lcccccccccccc}
\hlineB{2.5}
\centering
\multirow{2}{*}{\textbf{Method}} &
\multirow{2}{*}{\textbf{Paradigm}}
& \multicolumn{2}{c}{\textbf{Bonn}} 
& \multicolumn{2}{c}{\textbf{ScanNet}} 
& \multicolumn{2}{c}{\textbf{KITTI}} \\ 
\cmidrule(lr){3-4}\cmidrule(lr){5-6}\cmidrule(lr){7-8}
& & \textbf{AbsRel $\downarrow$} & \textbf{$\delta_1$ $\uparrow$} 
& \textbf{AbsRel $\downarrow$} & \textbf{$\delta_1$ $\uparrow$}
& \textbf{AbsRel $\downarrow$}  & \textbf{$\delta_1$ $\uparrow$}  \\
\hlineB{1.5}

DAv2-L~\citep{yang2024depthv2}& Diff.+D
& {8.7} & {0.952} 
& {9.5} & {0.940} 
& {11.9} & {0.879} \\
Marigold v1.1~\citep{ke2025marigold}& Diff.+G
& {11.6} & {0.890} 
& {12.0} & {0.870} 
& {24.7} & {0.569} \\
RollingDepth~\citep{ke2024rollingdepth}& Diff.+G
& {7.2} & {0.966} 
& \underline{7.5} & {0.957} 
& {11.1} & {0.911} \\
 \hline
ChronoDepth \citep{shao2025learning} & Diff.+G
& {17.3} & {0.859} 
& {21.2} & {0.715} 
& {13.0} & {0.846} \\
DepthCrafter \citep{hu2025depthcrafter}   & Diff.+G 
& {8.5} & {0.962} 
& {11.4} & {0.866} 
& {12.0} & {0.858} \\
VDA \citep{chen2025video}  & ViT+D 
& \underline{6.6} & \underline{0.971} 
& \textbf{7.3} & \underline{0.972} 
& \underline{9.6} & \underline{0.940} \\

\hline
\rowcolor{CadetBlue!20} 
\textbf{\ourmtab~(Ours)}  & Diff.+D
& \textbf{5.3} & \textbf{0.978} 
& \textbf{7.3} & \textbf{0.977} 
& \textbf{7.6} & \textbf{0.956} \\
\hlineB{2.5}
\end{tabular}
\label{tab:long_video}
\vspace{-0.8em}
\end{table*}

\subsection{Image-Video Joint Training}

Training exclusively on video data often compromises spatial sharpness, whereas sequential fine-tuning (image $\to$ video) may risk catastrophic forgetting of per-frame details. To bypass this trade-off, we optimize \ourmethod via an \textit{image-video joint training} strategy. By constructing batches comprising both static images ($F=1$) and dynamic video sequences, the images act as high-frequency spatial anchors while the videos enforce temporal coherence. The unified objective is simply formulated as:
\setlength\abovedisplayskip{3pt}
\setlength\belowdisplayskip{3pt}
\begin{equation}
    \mathcal{L}_\text{joint} = \mathcal{L}_\text{video} + \lambda_\text{image}\mathcal{L}_\text{image}.
\end{equation}
This simple yet effective strategy enables \ourmethod to maintain the spatial quality of diffusion priors while achieving robust temporal stability.

\section{Experiments}
\vspace{-0.2em}

\subsection{Experimental Setup}
\label{sec:exp_setup}

\textbf{Implementation Details.}\quad We adopt WanV2.1-1.3B~\citep{wan2025wan} as our \ourmethod's backbone, fine-tuned via LoRA following~\citep{he2025lotus}. We employ a joint image-video training strategy strictly using public synthetic datasets: video clips from TartanAir~\citep{tartanair2020iros} and Virtual KITTI~\citep{gaidon2016virtual} (batch size $16$), alongside static images from Hypersim~\citep{roberts2021hypersim} and Virtual KITTI (batch size $128$).
The entire framework converges in under $36$ hours on $8$ H$100$ GPUs, demonstrating higher training efficiency and eco-friendliness than prior arts~\citep{chen2025video,hu2025depthcrafter}. More details can be found in Appendix~\S\ref{apx:impl}.

\noindent\textbf{Evaluation.}\quad To assess both temporal consistency and per-frame accuracy, we conduct comprehensive evaluations across two settings:
\ding{182} \textbf{video datasets:} KITTI \citep{Geiger2012CVPR}, ScanNet \citep{dai2017scannet}, Bonn \citep{palazzolo2019bonn}, and Sintel \citep{Butler2012sintel}; and
\ding{183} \textbf{image datasets:} KITTI \citep{Geiger2012CVPR}, DIODE \citep{diode_dataset}, and NYUv2 \citep{SilbermanECCV12nyuv2}.
We report standard metrics including absolute relative error (AbsRel) and threshold accuracy ($\delta_1$), as well as boundary metrics like boundary F1-Score~(B-F1) and boundary Recall~(B-Recall)~\citep{bochkovskii2024depthpro}.

\noindent\textbf{Baselines.}\quad 
We compare \ourmethod with representative state-of-the-art video depth estimation methods from two paradigms:
\ding{182} \textbf{generative-based:} ChoronDepth \citep{shao2025learning} and DepthCrafter \citep{hu2025depthcrafter};
\ding{183} \textbf{discriminative-based:} Video Depth Anything (VDA) \citep{chen2025video}; along with image-based Depth Anything V2 (DAv2) \citep{yang2024depthv2}, Marigold \citep{ke2025marigold}, and RollingDepth \citep{ke2024rollingdepth} for reference. 
Following standard protocols \citep{ke2025marigold,yang2024depthv2,chen2025video,hu2025depthcrafter}, "zero-shot" denotes direct evaluation on target benchmarks without any domain-specific fine-tuning.

\begingroup
\setlength{\tabcolsep}{12.8pt}
\begin{table*}[t!]
\vspace{0.6em}
\renewcommand{\arraystretch}{1.2}
\centering
\caption{\textbf{Zero-shot boundary metrics, \textit{i.e.}, Recall and F1.}
Higher values indicate sharper boundaries and finer details.
}
\vspace{-0.8em}
\footnotesize
\begin{tabular}{lccccccc}
\hlineB{2.5}
\multirow{2}{*}{\textbf{Method}} &
\multicolumn{2}{c}{\textbf{Bonn}} &
\multicolumn{2}{c}{\textbf{ScanNet}} &
\multicolumn{2}{c}{\textbf{KITTI}} \\
\cmidrule(r){2-3}\cmidrule(r){4-5}\cmidrule(r){6-7}
& \textbf{B-Recall $\uparrow$} & \textbf{B-F1 $\uparrow$} & \textbf{B-Recall $\uparrow$} & \textbf{B-F1 $\uparrow$} & \textbf{B-Recall $\uparrow$} & \textbf{B-F1 $\uparrow$} \\
\hlineB{1.5}




ChronoDepth \citep{shao2025learning}
&  {0.221} & {0.319}
&  {0.144} & {0.204}
&  {0.049} & \underline{0.090} \\

DepthCrafter \citep{hu2025depthcrafter}
&  \underline{0.282} & {0.185}
&  {0.115} & {0.173}
&  \underline{0.082} & {0.044} \\

VDA \citep{chen2025video}
&  {0.223} & \underline{0.325}
&  \underline{0.147} & \underline{0.210}
&  {0.047} & {0.088} \\

\hline
\rowcolor{CadetBlue!20} 
\textbf{\ourmtab (Ours)} 
&  \textbf{0.336} & \textbf{0.422}
&  \textbf{0.208} & \textbf{0.259} 
&  \textbf{0.217} & \textbf{0.285}
\\

\hlineB{2.5}
\end{tabular}
\label{tab:boundary}
\vspace{-0.8em}
\end{table*}

\begingroup
\setlength{\tabcolsep}{14.1pt}
\begin{table*}[t!]
\renewcommand{\arraystretch}{1.2}
\centering
\caption{\textbf{Zero-shot single-image depth estimation results.}}
\vspace{-0.8em}
\footnotesize
\begin{tabular}{lccccccccccc}
\hlineB{2.5}
\centering
\multirow{2}{*}{\textbf{Method}} 
& \multicolumn{2}{c}{\textbf{KITTI}} 
& \multicolumn{2}{c}{\textbf{DIODE}} 
& \multicolumn{2}{c}{\textbf{NYUv2}} 
\\ 
\cmidrule(r){2-3}\cmidrule(r){4-5}\cmidrule(r){6-7}
& \textbf{AbsRel $\downarrow$} & \textbf{$\delta_1$ $\uparrow$} 
& \textbf{AbsRel $\downarrow$} & \textbf{$\delta_1$ $\uparrow$}
& \textbf{AbsRel $\downarrow$}  & \textbf{$\delta_1$ $\uparrow$}
\\
 \hlineB{1.5}


ChronoDepth \citep{shao2025learning}    
& {-} & {-} 
& {80.3} & {0.549} 
& {21.2}    & {0.767}    
\\

DepthCrafter \citep{hu2025depthcrafter}  
& {11.0} & {0.877} 
& {53.3} & {0.592} 
& {17.1}    & {0.868}    
\\

VDA \citep{chen2025video}      
& \underline{8.3} & \underline{0.933} 
& \underline{27.0} & \underline{0.730} 
& \textbf{4.7}    & \textbf{0.977}    
\\

\hline
\rowcolor{CadetBlue!20} 
\textbf{\ourmtab~(Ours)}  
& \textbf{8.1} & \textbf{0.944} 
& \textbf{23.1} & \textbf{0.738} 
& \underline{5.5}    & \underline{0.969}   
\\
 \hlineB{2.5}
\end{tabular}
\label{tab:image}
\vspace{-0.8em}
\end{table*}

\subsection{Main Results}
\label{sec:main_results}

This section provides empirical evidence of \ourmethod's effectiveness. We evaluate our approach across standard video depth benchmarks (Table~\ref{tab:short_video}), long-term consistency tasks (Table~\ref{tab:long_video}), fine-grained boundary metrics (Table~\ref{tab:boundary}), and single-image generalization (Table~\ref{tab:image}). Qualitative comparisons (Figure~\ref{fig:teaser}, \ref{fig:qualitative} and \ref{fig:qualitative_long}) and rigorous efficiency \& scalability analyses (Figure~\ref{fig:main_results}) further validate our design. Our key observations are summarized as follows:

\begin{figure}[!t]
    \centering
    \includegraphics[width=\linewidth]{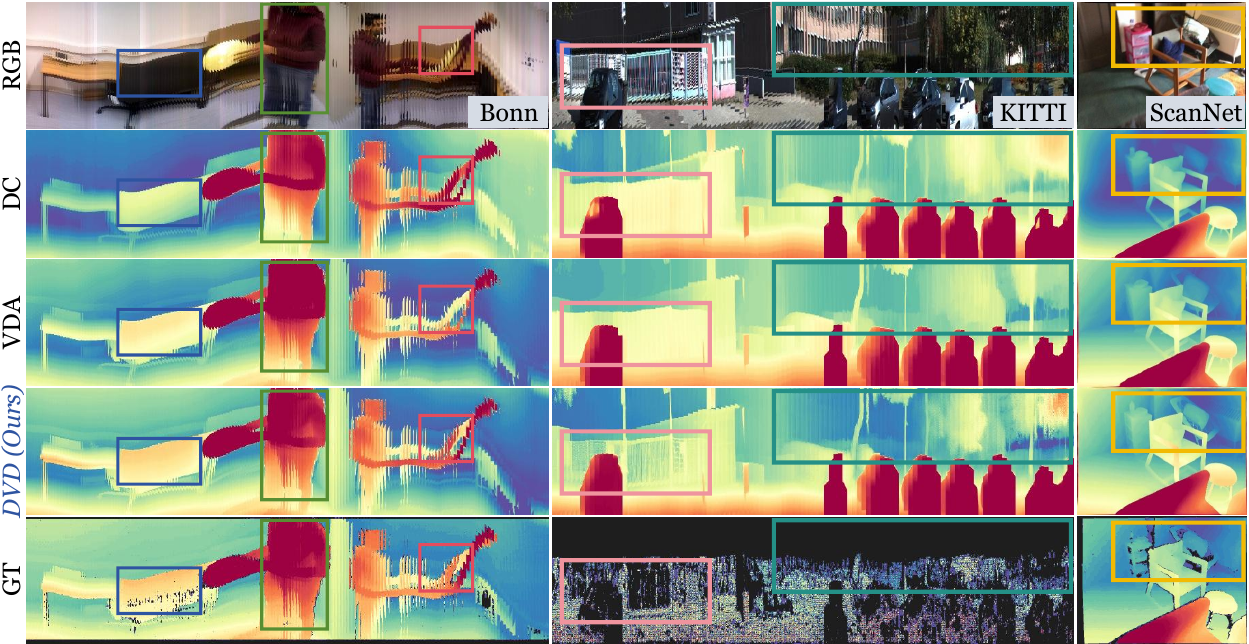}
    \vspace{-2em}
    \caption{
     \textbf{Qualitative comparison on indoor and outdoor scenes.}
    \ourmethod consistently produces higher fidelity depth with noticeably sharper structural boundaries
    }
    \label{fig:qualitative}
    \vspace{-1em}
\end{figure}

\begin{figure*}[!t]
\centering
\vspace{0.4em}
\includegraphics[width=\linewidth]{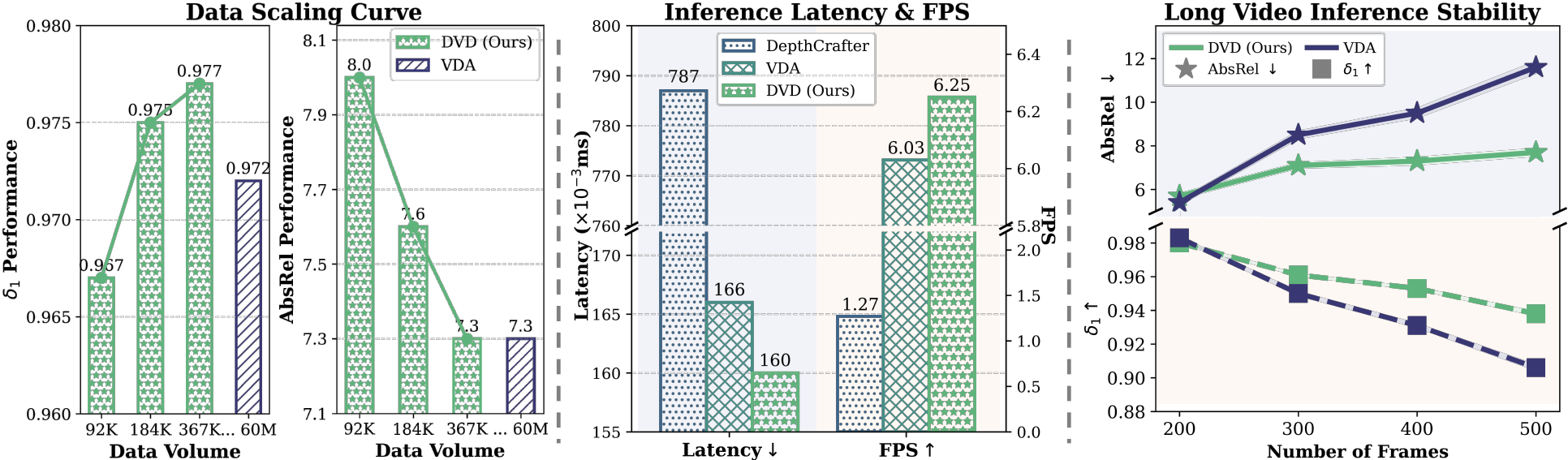}
\vspace{-2em}
\caption{(\textbf{\textit{Left}}) Data scaling curve of long video performance on ScanNet. (\textbf{\textit{Middle}}) Inference latency \& FPS comparisons. (\textbf{\textit{Right}}) Stability of long video inference on KITTI. These results demonstrate \ourmethod's remarkable data efficiency, competitive inference speed, and consistent long-video stability.}
\label{fig:main_results}
\vspace{-0.8em}
\end{figure*}

\textbf{Obs.\ding{182} \ourmethod achieves superior geometric fidelity and temporal coherence.} 
Across standard real world benchmarks (Table~\ref{tab:short_video}), \ourmethod consistently outperforms state-of-the-art generative (\textit{e.g.}, DepthCrafter) and discriminative (\textit{e.g.}, VDA) baselines, achieving the lowest AbsRel on ScanNet ($5.5$) and KITTI ($6.7$). This superiority extends to long-video scenarios (Table~\ref{tab:long_video}), yielding a substantial margin over DepthCrafter on Bonn ($5.3$ \textit{vs.} $8.5$ AbsRel). Beyond global metrics, \ourmethod excels at preserving fine-grained geometry. Table~\ref{tab:boundary} and Figure~\ref{fig:qualitative} demonstrate that our latent manifold rectification successfully combats mean collapse, significantly boosting the ScanNet B-F1 score to $0.259$ (compared to VDA's $0.210$). Importantly, our joint training strategy ensures this temporal robustness does not compromise spatial precision, retaining highly competitive single-image generalization, as shown in Table~\ref{tab:image}.

\textbf{Obs.\ding{183} \ourmethod exhibits compelling data and inference efficiency.}
A pivotal advantage of \ourmethod is unlocking high-fidelity depth with minimal task-specific data and low latency. As illustrated in the scaling curves in Figure~\ref{fig:main_results} (\textit{Left}) and Table~\ref{tab:short_video}, our model trained on just $367$K frames surpasses VDA, utilizing less than $1/160$ of its massive dataset ($60$M frames). This confirms that deterministic adaptation of pre-trained world models is a vastly more data-efficient paradigm. Furthermore, as shown in the inference latency analysis in Figure~\ref{fig:main_results} (\textit{Middle}), \ourmethod completely bypasses the computational bottleneck of iterative generative sampling, maintaining an inference speed comparable to efficient discriminative models like VDA while delivering superior accuracy.

\textbf{Obs.\ding{184} \ourmethod exhibits robust scalability to long videos.}
Beyond short clips, \ourmethod maintains inherent global scale consistency across disjoint temporal windows. As visualized in the in-the-wild (Figure~\ref{fig:teaser}) and complex domestic (Figure~\ref{fig:qualitative_long}) sequences long-video sequences, while generative methods (\textit{e.g.}, DepthCrafter) suffer from severe scale drift and discriminative baselines (\textit{e.g.}, VDA) persistently exhibit semantic ambiguity, our parameter-free affine-alignment mechanism ensures strict structural persistence and high fidelity over thousands of frames. This is quantitatively validated in Figure~\ref{fig:main_results} (\textit{Right}): as the sequence length increases, baseline methods exhibit more pronounced metric degradation, whereas \ourmethod maintains more consistent stability. More qualitative analyses across diverse open-world scenes are provided in Appendix~\S\ref{apx:viz}.

\begin{figure}[!t]
    \centering
    \includegraphics[width=\linewidth]{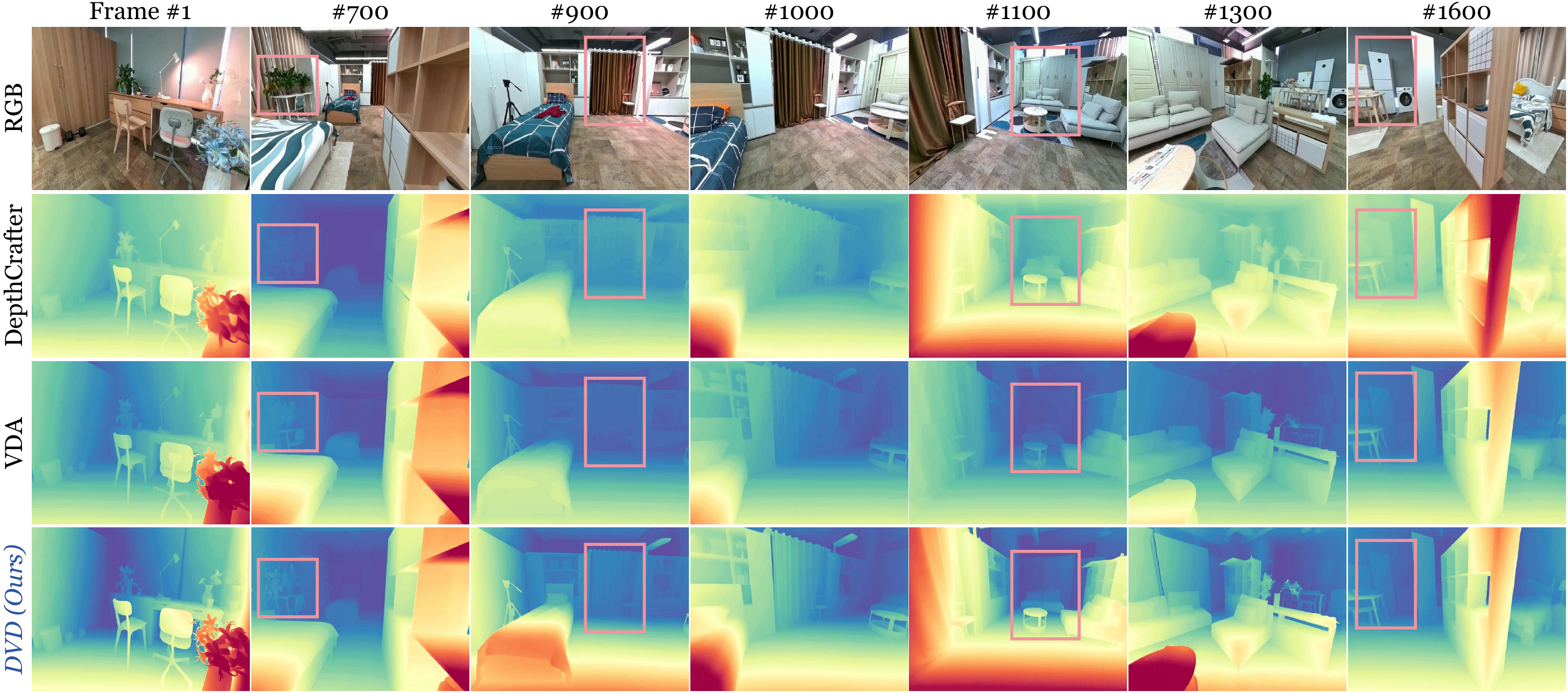}
    \vspace{-2em}
    \caption{
     \textbf{Qualitative results on long-horizon indoor navigation.} \ourmethod leverages global affine coherence to better preserve sharp boundaries and globally coherent geometry across thousands of frames compared to prior SOTA methods.
    }
    \label{fig:qualitative_long}
    \vspace{-0.8em}
\end{figure}

\subsection{Framework Analysis}
\label{sec:5.3}

In this section, we analyze the core design choices of \ourmethod. Unless otherwise specified, all ablation experiments are conducted on ScanNet~\citep{dai2017scannet}. More studies are provided in Appendix~\S\ref{apx:more_ablation}.

\begin{wrapfigure}{r}{0.46\textwidth}
\vspace{-1.7em}
 \centering
 \includegraphics[width=\linewidth]{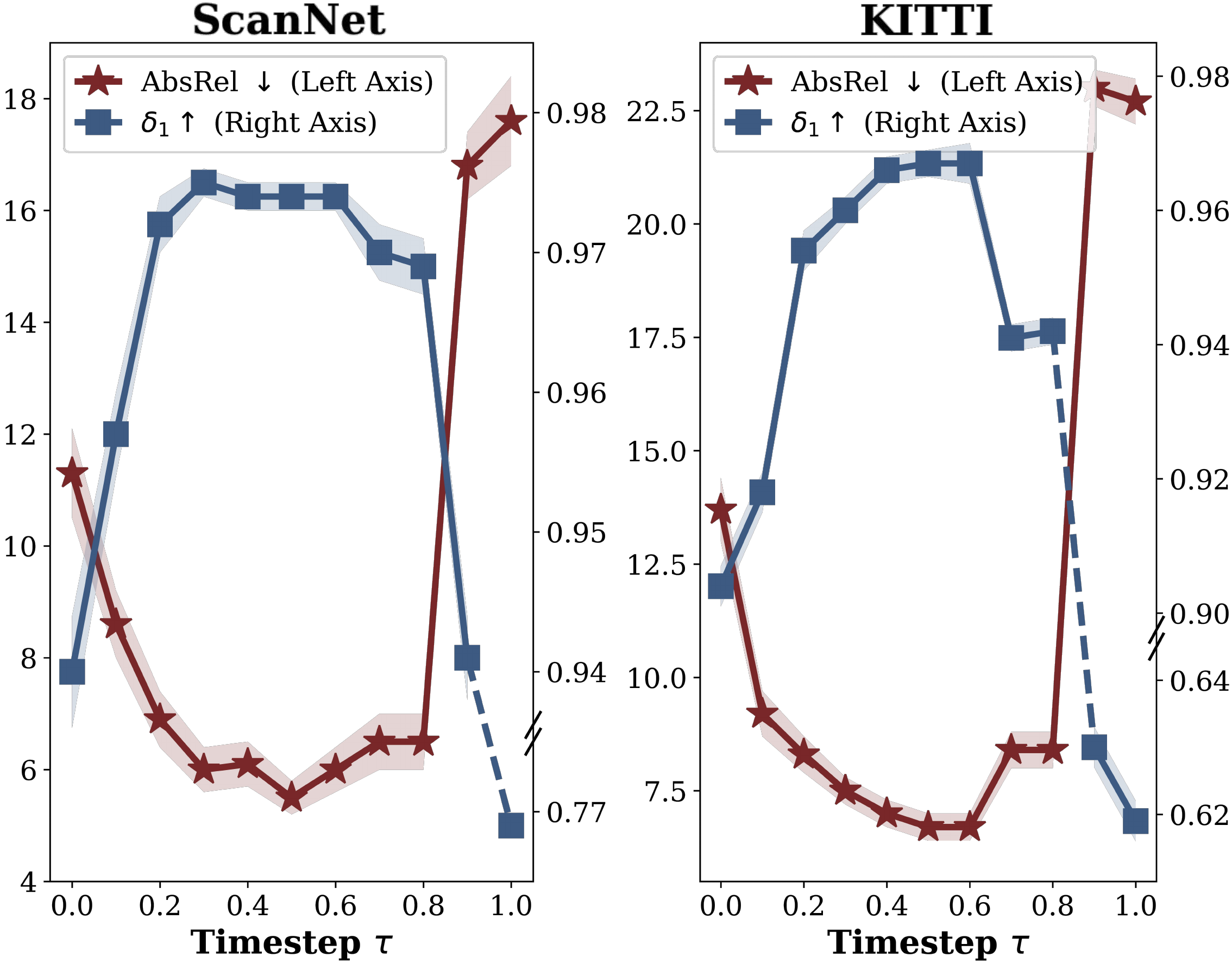}
  \vspace{-2.1em}
  \caption{\textbf{Ablation of timestep $\tau$.} The structural anchor $\tau$ dictates a fidelity-stability trade-off. While outdoor scenes are more sensitive, both datasets achieve a balance at $\tau = 0.5$.}
  \vspace{-2.4em}
  \label{fig:ablation_timestep}
\end{wrapfigure}
\noindent\textbf{Role of Timestep Conditioning.}\quad
We first investigate the impact of the structural anchor $\tau$ in Figure~\ref{fig:ablation_timestep}. The model exhibits a clear fidelity-stability trade-off, where $\tau=0.5$ achieves the optimal balance. Notably, comparing indoor (\textit{Left}) and outdoor (\textit{Right}) scenes reveals that outdoor environments are significantly more sensitive to the structural anchor, with KITTI's AbsRel fluctuating drastically from $13.8$ ($\tau=0.0$) down to $6.7$ ($\tau=0.5$). Furthermore, pushing the anchor to extreme high values ($\tau \ge 0.9$) triggers a severe performance collapse across both datasets, as the extreme low-frequency bias completely washes out essential geometric details. This confirms that $\tau$ acts as a persistent structural anchor dictating the pre-trained geometric operating regime.

\noindent\textbf{Impact of Latent Manifold Rectification.}\quad
We evaluate our gradient-aware LMR modules in Figure~\ref{fig:ablation} (\textit{Left}). Adding the spatial latent gradient ($\mathcal{L}_\text{sp}$) and temporal latent flow ($\mathcal{L}_\text{temp}$) progressively improves both global accuracy (AbsRel drops from $8.5$ to $7.3$) and fine-grained boundary precision (B-F1 rises from $0.210$ to $0.259$). This proves that explicitly enforcing differential constraints successfully rectifies the mean collapse inherent in single-pass regression, restoring sharp structural boundaries and coherent motion. Comparisons with alternative regularizations are placed in  Appendix~\S\ref{apx:more_ablation}.

\noindent\textbf{Deterministic Adaptation \textit{vs.} Stochastic Sampling.}\quad
We compare our single-pass deterministic regression against standard multi-step sampling in Figure~\ref{fig:ablation} (\textit{Middle}). Our deterministic adaptation significantly outperforms generative multi-step sampling in geometric accuracy, dropping AbsRel from $9.7$ ($T=10$) to $7.3$ ($T=1$). This validates our core hypothesis: iterative stochastic sampling introduces aleatoric variance that manifests as geometric hallucinations, whereas our deterministic regression directly targets the conditional geometric expectation, ensuring superior structural stability.

\begin{figure*}[!t]
\centering
\includegraphics[width=\linewidth]{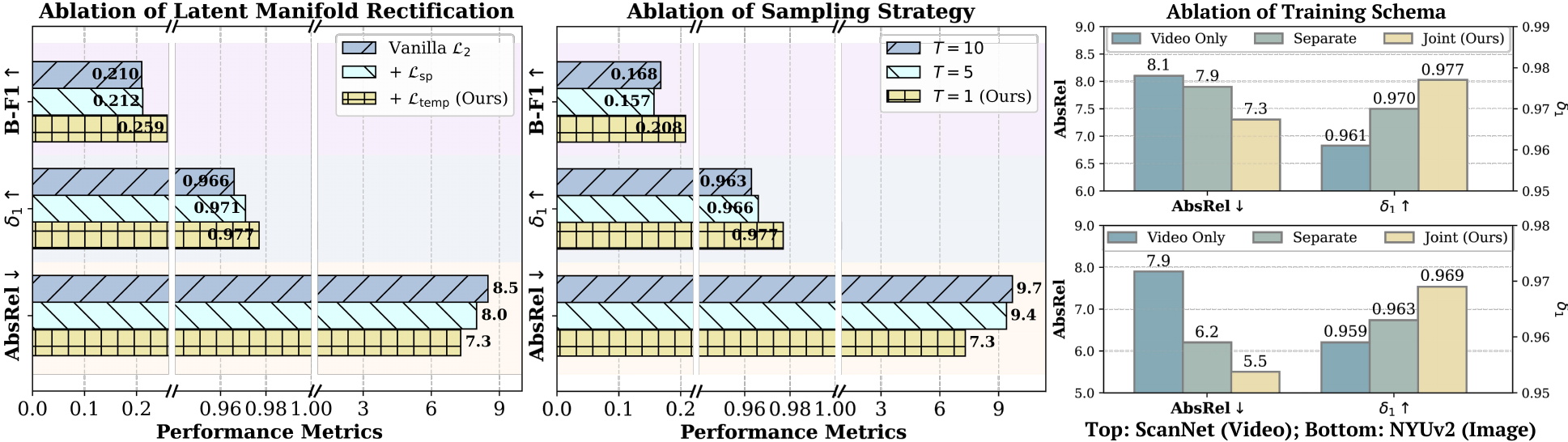}
\vspace{-1.8em}
\caption{(\textbf{\textit{Left}}) Ablation of latent manifold rectification. (\textbf{\textit{Middle}}) Ablation of the sampling strategy. (\textbf{\textit{Right}}) Ablation of training schema. The results demonstrate that latent manifold rectification, deterministic adaptation, and joint image-video training effectively enhance geometric accuracy and structural stability.}
\label{fig:ablation}
\vspace{-0.8em}
\end{figure*}

\noindent\textbf{Effectiveness of Image-Video Joint Training.}\quad
Finally, Figure~\ref{fig:ablation} (\textit{Right}) analyzes our training strategy. Training exclusively on videos underfits spatial details, while separate sequential training suffers from catastrophic forgetting on single-frame tasks. Our joint strategy achieves the best performance across both domains, \textit{i.e.}, maximizing ScanNet video accuracy ($\delta_1 = 0.977$) while sharply reducing NYUv2 single-image AbsRel to $5.5$. This demonstrates that images and videos offer complementary supervision: images act as high-frequency spatial anchors, while videos enforce temporal consistency.

\section{Conclusion}

In this work, we present \ourmethod, the first framework to deterministically adapt pre-trained video diffusion priors for single-pass depth estimation. By bypassing stochastic sampling, \ourmethod successfully resolves the ambiguity-hallucination dilemma, uniting the semantic richness of generative models with the structural stability of discriminative regressors. Our three core designs, (\textbf{\textit{i}}) a timestep-driven structural anchor, (\textbf{\textit{ii}}) latent manifold rectification (LMR) against spatio-temporal mean collapse, and (\textbf{\textit{iii}}) global affine coherence for affine-alignment long-video inference, collectively establish a robust zero-shot solution. Crucially, by effectively grounding these generative priors, \ourmethod achieves state-of-the-art geometric fidelity and temporal coherence while utilizing 163$\times$ less task-specific training data than leading baselines. This establishes a highly scalable and data-efficient paradigm for dynamic 3D scene understanding.

\newpage
\bibliographystyle{assets/plainnat}
\bibliography{paper}

\clearpage
\newpage
\beginappendix

\section{Limitations and Future Work}
\label{apx:limitations}
While \ourmethod establishes a robust paradigm for video depth estimation, we acknowledge several limitations that highlight promising avenues for future research.

\noindent\textbf{Boundary Conditions of Long Videos.}\quad 
In unconstrained long videos, extreme dynamics, such as prolonged occlusions, rapid illumination shifts, or erratic camera motions, can introduce local non-linear distortions that temporarily overpower our global affine assumption, leading to scale inconsistencies. Future work could explore larger temporal context windows or non-linear latent tracking to mitigate these edge cases. Detailed visual failure analyses are provided in Section~\S\ref{app:failure}.

\noindent\textbf{Constraints for Real-Time Deployment.}\quad 
Although bypassing stochastic sampling significantly accelerates inference, \ourmethod still relies on a massive video DiT backbone (\textit{e.g.}, Wan2.1-1.3B~\citep{wan2025wan}). Achieving true real-time inference (\textit{e.g.}, $\ge$ 10Hz) for latency-critical on-device applications remains challenging. Promising future directions include architectural distillation and integrating efficient linear-complexity sequence models to transfer these profound generative priors into lightweight networks.

\noindent\textbf{Resolution Limits of VAE.}\quad 
Operating within a highly compressed VAE latent space (\textit{e.g.}, $8\times$ downsampling) inherently upper-bounds the recovery of ultra-thin geometric structures at native resolutions. While our latent manifold rectification (LMR) effectively mitigates structural collapse, transitioning to higher-resolution latent spaces or fully VAE-free tokenization schemes presents a compelling direction to further push geometric fidelity.


%

\begin{figure}[!b]
    \centering
    \includegraphics[width=\linewidth]{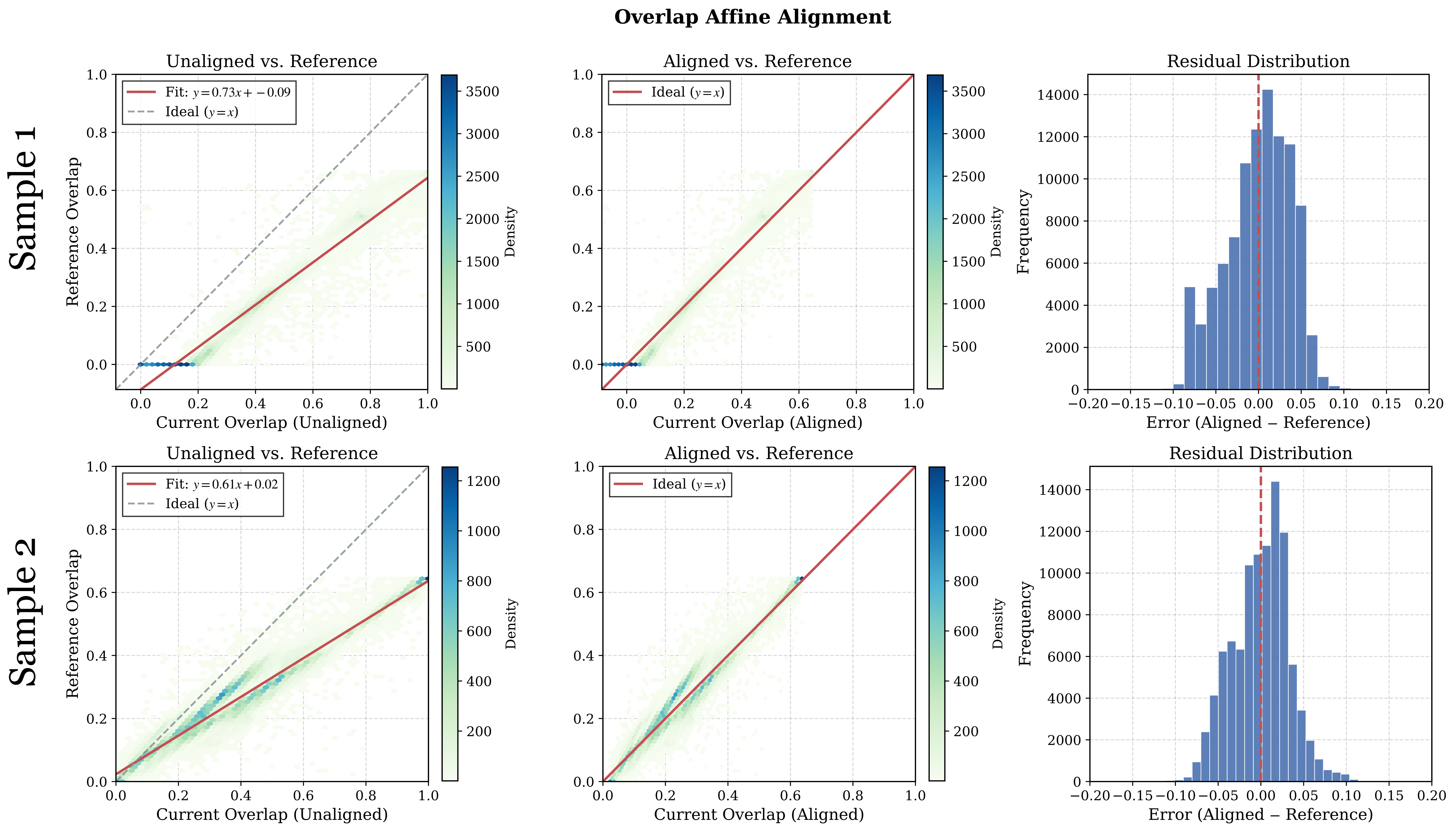}
    \vspace{-2em}
    \caption{
    Visualization of inter-window affine alignment.
    (\textbf{\textit{Left}}) Joint pixel density distribution between the current and reference windows in the overlapping region. The strictly linear correlation (red line) confirms that the inter-window discrepancy is predominantly affine. (\textbf{\textit{Middle}}) After applying our calculated affine transformation, the predictions tightly cluster along the ideal $y=x$ diagonal. (\textbf{\textit{Right}}) The histogram of pixel-wise residuals (Aligned minus Reference) exhibits a zero-mean, tightly bounded distribution with minimal variance, demonstrating the high precision of our global affine coherence strategy.
    }
    \label{app_fig:gac_study}
\end{figure}

\section{More Details of Global Affine Coherence}

To further validate the core assumption underpinning our global affine coherence, we visualize the pixel-wise depth relationship within overlapping temporal windows here. As shown in Figure~\ref{app_fig:gac_study} (\textit{Left}), the joint density distribution of unaligned predictions forms a strictly linear trajectory. This striking visual evidence explicitly corroborates our hypothesis: inter-window discrepancies, primarily induced by the deterministic VAE decoding, are fundamentally dominated by global scale and shift factors, rather than complex non-linear distortions. By applying our calculated affine transformation, the predictions perfectly converge onto the ideal $y=x$ diagonal (\textit{Middle}). Furthermore, the post-alignment residual distribution (\textit{Right}) is strictly zero-centered with minimal variance. This demonstrates that our lightweight linear alignment strategy is effective for stitching long sequences with negligible geometric error.

\section{More Implementation Details}
\label{apx:impl}

\textbf{Training Details.}\quad To facilitate reproducibility, we detail the architectural configurations and training hyperparameters of \ourmethod in this section. To adapt this model for deterministic regression while strictly preserving its pre-trained world priors, we freeze the original weights and employ Low-Rank Adaptation (LoRA) exclusively on the attention blocks. The detailed settings for the VAE compression, LoRA configuration, optimization schedule, and joint-training loss weights are systematically summarized in Table~\ref{tab:hyperparameters}.

\begingroup
\setlength{\tabcolsep}{36pt}
\begin{table*}[!h]
\vspace{-0.2em}
\renewcommand{\arraystretch}{1.16}
\caption{Hyperparameter configurations for \ourmethod.}
\vspace{-1em}
\label{tab:hyperparameters}
\centering
\footnotesize
\begin{tabular}{llc}
\hlineB{2.5}
\rowcolor{CadetBlue!20} 
\textbf{Setting} & \textbf{Hyperparameter} & \textbf{Value} \\
\hlineB{1.5}
\multicolumn{3}{c}{\textit{Model Architecture $\&$ Adaptation}} \\
\hline
\rowcolor{gray!10}
\textbf{Backbone} & pre-trained model & Wan2.1-1.3B \\
\textbf{VAE} & spatial compression & $8\times$ \\
& temporal compression & $4\times$ \\
\rowcolor{gray!10}
\textbf{LoRA} & target modules & $W_q,W_k,W_v,W_o,W_{ffn}$ \\
\rowcolor{gray!10}
& rank ($r$) / alpha ($\alpha$) & $512/512$ \\
\hlineB{1.5}
\multicolumn{3}{c}{\textit{Training $\&$ Inference Settings}} \\
\hline
\rowcolor{gray!10}
\textbf{Optimization} & optimizer & AdamW \\
\rowcolor{gray!10}
& base learning rate & $1\times10^{-4}$ \\
\rowcolor{gray!10}
& LR schedule & Constant \\
\rowcolor{gray!10}
& hardware & $8\times$ NVIDIA H100 GPUs \\
\textbf{Data} & global batch size (video) & $16$ \\
& global batch size (image) & $128$ \\
& spatial resolution & $480\times640$ \\
& window size & $45$ \\
& stride (inference) & $9$ \\
\hlineB{1.5}
\multicolumn{3}{c}{\textit{Objective Weights}} \\
\hline
\rowcolor{gray!10}
\textbf{Loss Weights} & spatial rectification ($\lambda_\text{sp}$) & $0.5$ \\
\rowcolor{gray!10}
& temporal rectification ($\lambda_\text{temp}$) & $0.5$ \\
\rowcolor{gray!10}
& image joint loss ($\lambda_\text{image}$) & $1.0$ \\
\hlineB{2}
\end{tabular}
\vspace{-0.4em}
\end{table*} 
\endgroup

\noindent\textbf{Experimental Details.}\quad 
We provide more details here regarding our fine-grained evaluation protocols and inference efficiency benchmarking to ensure fair comparison and reproducibility. \ding{182} \textbf{Video Length Partitioning.} Aggregating metrics across an entire dataset often masks the severe scale drift generative models suffer on extended sequences. To rigorously assess temporal stability, we partition the test sets by duration: \textit{short videos} ($50$--$200$ frames) and \textit{long videos} ($>200$ frames). This explicit decoupling effectively isolates short-term geometric fidelity from long-term structural persistence. \ding{183} \textbf{Inference Efficiency Benchmarking.} Latency and FPS are evaluated on a single NVIDIA RTX A6000 GPU under identical environments. To reflect practical deployment, we implement two streamlined optimizations: (\textit{\textbf{i}}) merging LoRA weights into the base backbone to eliminate modular overhead, and (\textit{\textbf{ii}}) using non-tiled VAE decoding to bypass spatial slicing bottlenecks. While our single-pass paradigm is inherently fast, integrating advanced accelerators (\textit{e.g.}, TensorRT) remains a promising future direction for real-time edge applications.

\begingroup
\setlength{\tabcolsep}{21pt}
\begin{table*}[!t]
\renewcommand{\arraystretch}{1.2}
\centering
\caption{Cross-backbone generalization on KITTI. We deploy \ourmethod on CogVideoX-5B \citep{hong2022cogvideo}. Consistent with our default Wan2.1-1.3B backbone, the mid-range timestep ($\tau=0.5$) serves as the optimal structural anchor.}
\vspace{-1em}
\centering
\footnotesize
\begin{tabular}{c|c|ccccc}
\hlineB{2.5}
\rowcolor{CadetBlue!20} & \textbf{Wan2.1-1.3B} & \multicolumn{5}{c}{\textbf{CogVideoX-5B (by $\tau$)}} \\
\rowcolor{CadetBlue!20}
\multirow{-2}{*}{\textbf{Metric}} & \ourmethod & \textbf{0.1} & \textbf{0.3} & \textbf{0.5} & \textbf{0.7} & \textbf{0.9} \\
 \hlineB{1.5}
AbsRel $\downarrow$ & 6.7 & 7.8 & 7.6 & 7.4 & 7.4 & 9.5 \\
\rowcolor{gray!10}
$\delta \uparrow$ & 0.967 & 0.931 & 0.934 & 0.938 & 0.935 & 0.898 \\
 \hlineB{2.5}
\end{tabular}
\label{app_tab:ablation_backbone_kitti}
\vspace{-1em}
\end{table*}

\begin{figure}[!t]
    \centering
    \includegraphics[width=\linewidth]{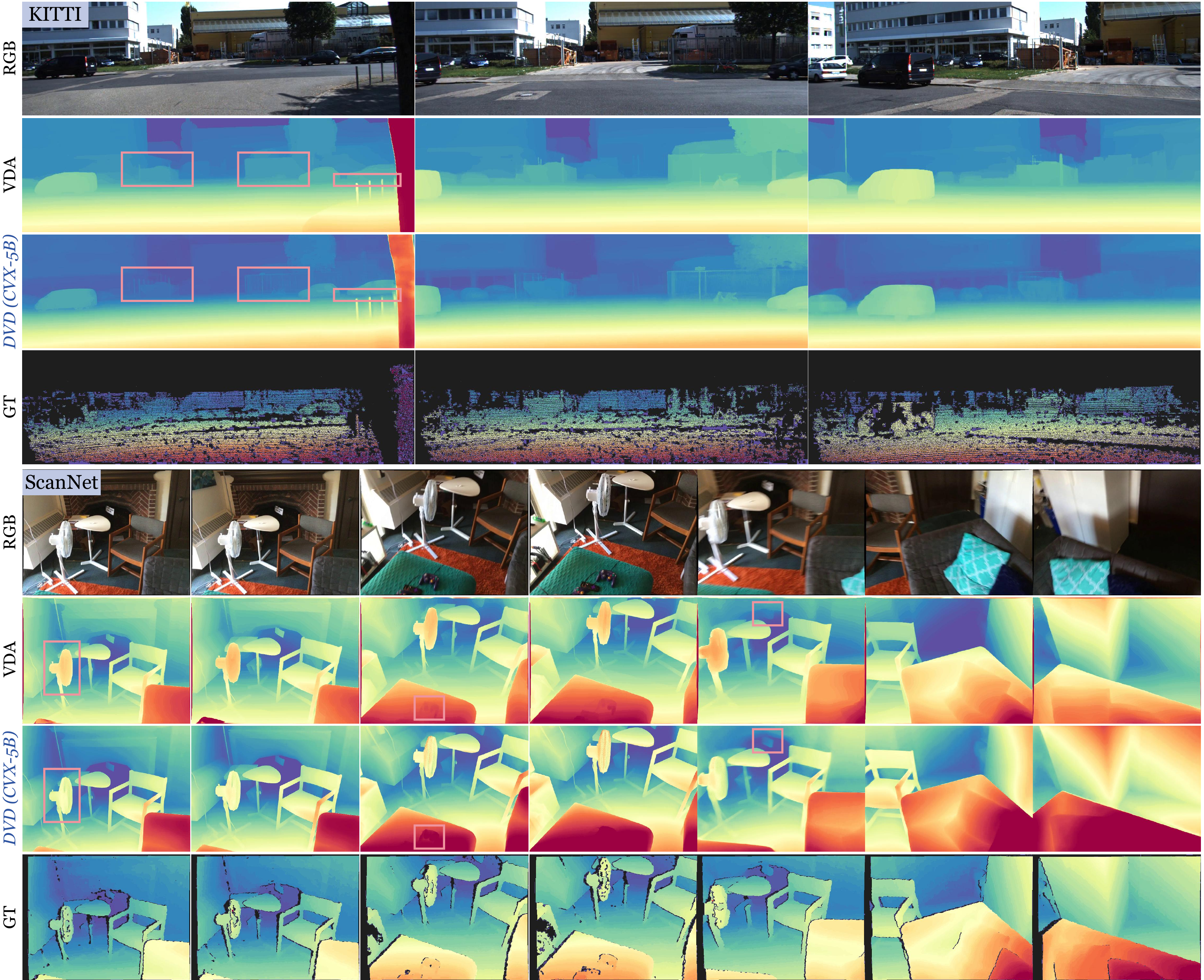}
    \vspace{-2em}
    \caption{Qualitative results of \ourmethod on CogVideoX~\citep{hong2022cogvideo}. Despite employing a different foundation architecture, our deterministic paradigm still preserves significantly sharper high-frequency details (highlighted in red boxes) compared to the leading baseline.
    }
    \label{app_fig:cogvideox}
    \vspace{-0.8em}
\end{figure}

\section{More Analysis}
\label{apx:more_ablation}

\noindent\textbf{Cross-Backbone Generalization.}\quad To verify that our deterministic adaptation paradigm is universally applicable rather than specific to a single architecture, we further deploy \ourmethod on CogVideoX-5B~\citep{hong2022cogvideo}. As shown in Table~\ref{app_tab:ablation_backbone_kitti}, ablating the structural anchor $\tau$ on KITTI perfectly corroborates our findings from the Wan2.1-1.3B backbone: extreme timesteps degrade geometry, whereas the mid-range value ($\tau=0.5$) provides the optimal conditioning (AbsRel $7.4$, $\delta_1$ $0.938$). 
While the CogVideoX-5B $+$\ourmethod variant yields expectedly lower metrics than our default Wan2.1 backbone (due to differing foundation capacities and pre-training distributions), its fundamental capability for structural extraction remains profoundly robust. As visualized in Figure~\ref{app_fig:cogvideox}, even with a different generative backbone, \ourmethod effortlessly recovers fine-grained, high-frequency geometries that are severely over-smoothed by the state-of-the-art discriminative baseline, VDA. This further confirms the broad generalizability of our zero-shot adaptation strategy across diverse generative families.

\begin{wraptable}{r}{0.4\textwidth}
 \vspace{-0.3em}
 \centering
    \caption{Analysis of structural anchor $\tau$, which is evaluated on ScanNet (\textbf{\textit{Left}}) and KITTI (\textbf{\textit{Right}}).}
    \vspace{-1em}
    \renewcommand\tabcolsep{5.8pt}
    \renewcommand\arraystretch{1.2}
    \footnotesize
    \label{app_tab:ablation_timestep}
    \begin{tabular}{lcccccc}
        \hlineB{2.5}
        \rowcolor{CadetBlue!20} 
        \textbf{Timestep} & \textbf{AbsRel$\downarrow$} & \textbf{$\delta_1\uparrow$} & \textbf{AbsRel$\downarrow$} & \textbf{$\delta_1\uparrow$} \\
        \hlineB{1.5}
        w/o $\tau$   & 11.3 & 0.940  &13.7& 0.904 \\
        \rowcolor{gray!10}
        $\tau=0.1$ & 8.6 & 0.957 & 9.2 & 0.918\\
        $\tau = 0.2$ & 6.9 & 0.972  & 8.3 & 0.954 \\
        \rowcolor{gray!10}
        $\tau = 0.3$ &  6.0 & 0.975  & 7.5 & 0.960 \\
        $\tau = 0.4$ & 6.1 & 0.974  & 7.0 & 0.966 \\
        \rowcolor{gray!10}
        $\tau=0.5$   & 5.5 & 0.974  & 6.7 & 0.967 \\
        $\tau = 0.6$ & 6.0 & 0.974  & 6.7 & 0.967 \\
        \rowcolor{gray!10}
        $\tau = 0.7$ & 6.5 & 0.970  & 8.4 & 0.941 \\
        $\tau = 0.8$ & 6.5 & 0.969  & 8.4 & 0.942 \\
        \rowcolor{gray!10}
        $\tau = 0.9$ & 16.8 & 0.941  & 23.0 & 0.630 \\
        $\tau = 1.0$ & 17.6 & 0.769  &22.7  & 0.619 \\
        \hline
        \rowcolor{gray!10}
        \textbf{learning $\tau$} &16.3&0.811&23.7&0.699\\
        \hlineB{2.5}
    \end{tabular}
    \vspace{-1.4em}
\end{wraptable}
\noindent\textbf{Analysis of Pre-trained Structural Anchors.}\quad
Table~\ref{app_tab:ablation_timestep} details the exact numerical breakdown of the fidelity-stability trade-off discussed in Figure~\ref{fig:ablation_timestep}. As previously established, entirely removing the structural anchor (represented as w/o $\tau$, which is equivalent to the $\tau=0.0$ boundary state) significantly degrades global metric accuracy. 
To further investigate whether this conditioning acts merely as a standard trainable parameter or an irreplaceable pre-trained key, we introduce a new extreme ablation: \textbf{learning $\tau$}. In this setting, we replace the fixed sinusoidal frequency basis with a randomly initialized, fully learnable embedding of identical dimensions. As shown in Table~\ref{app_tab:ablation_timestep}, learning a new anchor from scratch triggers a catastrophic performance collapse, with AbsRel skyrocketing to $16.3$ and $23.7$ on ScanNet and KITTI, respectively. We attribute this to the fact that the profound structural priors within the pre-trained video DiT backbone are fundamentally entangled with its original sinusoidal frequency encodings. A newly initialized embedding fails to activate these pre-trained pathways, effectively rendering the zero-shot generative priors inaccessible. This confirms that our fixed $\tau$ anchor natively unlocks the foundation model's geometric capacity, and cannot be replaced by naive fine-tuning.

\begin{wraptable}{r}{0.5\textwidth}
 \vspace{-1.4em}
 \centering
    \caption{Analysis of regularization strategies. All variants use the same backbone, training, and inference settings.}
    \vspace{-1em}
    \renewcommand\tabcolsep{5.8pt}
    \renewcommand\arraystretch{1.2}
    \footnotesize
    \label{app_tab:lmr_alt}
    \begin{tabular}{lcccc}
        \hlineB{2.5}
        \rowcolor{CadetBlue!20} 
        \textbf{Regularizer} & \textbf{AbsRel}$\downarrow$ & $\boldsymbol{\delta_1}\uparrow$ & \textbf{B-F1}$\uparrow$ \\
        \hlineB{1.5}
        $\mathcal{L}_2$ only (Eq.(\ref{eq:L2}))                & 8.5 & 0.966 & 0.210 \\
        \rowcolor{gray!10}
        $+$ RGB reconstruction & 10.5 & 0.951 & 0.174 \\
        $+$ Edge-aware smoothness         & \underline{7.5}  & \textbf{0.978} & 0.193 \\
        \rowcolor{gray!10}
        $+$ Multi-scale gradient matching     & 8.2  & 0.969 & \underline{0.257} \\
        \hline
        \textbf{$+$ LMR (Ours)}   & \textbf{7.3} &\underline{0.977} & \textbf{0.259} \\
        \hlineB{2.5}
    \end{tabular}
    \vspace{-1.4em}
\end{wraptable}
\noindent\textbf{Analysis of Different Regularization.}\quad
To isolate Latent Manifold Rectification (LMR)'s efficacy against mean collapse, we compare it with widely adopted regularizers in Table~\ref{app_tab:lmr_alt}. The vanilla $\mathcal{L}_2$ baseline yields sub-optimal accuracy (AbsRel $8.5$) and structural fidelity (B-F1 $0.210$). Adding RGB reconstruction distracts the network toward decoding textures rather than geometry. Interestingly, existing geometric regularizers present a strict trade-off: edge-aware smoothness improves global metrics (AbsRel $7.5$, $\delta_1$ $0.978$) but severely over-smoothes high-frequency details (B-F1 plummets to $0.193$). Conversely, multi-scale gradient matching sharpens boundaries (B-F1 $0.257$) but offers marginal global scale correction (AbsRel $8.2$). In stark contrast, LMR breaks this dilemma. By enforcing latent differential constraints, it simultaneously minimizes AbsRel ($7.3$) and maximizes boundary precision (B-F1 $0.259$), confirming its absolute superiority for deterministic adaptations.

\begin{wraptable}{r}{0.37\textwidth}
 \vspace{-1.4em}
 \centering
    \caption{Analysis of overlap size on KITTI. Increasing the number of overlapping frames ($O$) improves geometric accuracy but incurs diminishing returns and computational overhead (Rel. Time).}
    \vspace{-1em}
    \renewcommand\tabcolsep{5.8pt}
    \renewcommand\arraystretch{1.2}
    \footnotesize
    \label{app_tab:overlap}
    \begin{tabular}{lccc}
        \hlineB{2.5}
        \rowcolor{CadetBlue!20} 
        \textbf{Overlap Size} & \textbf{AbsRel}$\downarrow$ & $\boldsymbol{\delta_1}\uparrow$ & \textbf{Rel. Time}$\downarrow$ \\
        \hlineB{1.5}
        $O = 3$  & 7.9 & 0.937 & 1.00$\times$ \\
        \rowcolor{gray!10}
        $O = 6$  & 7.7 & 0.941 & 1.04$\times$ \\
        $O = 9$  & 7.3 & 0.945 & 1.17$\times$ \\
        \rowcolor{gray!10}
        $O = 14$  & {7.2} & {0.948} & {1.34}$\times$ \\
        $O = 19$  & {7.1} & {0.947} & {1.55}$\times$ \\
        \hlineB{2.5}
    \end{tabular}
    \vspace{-1.4em}
\end{wraptable}

\noindent\textbf{Analysis of Overlap Size.}\quad 
To determine the optimal balance between temporal consistency and computational efficiency, we further ablate the sliding window overlap size ($O$) on KITTI~\citep{Geiger2012CVPR} here. As detailed in Table~\ref{app_tab:overlap}, an extremely small overlap ($O=3$) yields sub-optimal accuracy (AbsRel $7.9$), as the limited pixel population makes the affine estimation susceptible to local dynamic outliers. Expanding the overlap region enriches the statistical basis for our Global Affine Coherence, effectively reducing inter-window discrepancies (\textit{e.g.}, AbsRel drops to $7.3$ at $O=9$). However, further enlarging the overlap ($O \ge 14$) yields diminishing geometric returns, and even saturates structural fidelity ($\delta_1$ slightly drops at $O=19$), while incurring severe computational overhead (reaching $1.55\times$ relative latency). Consequently, a moderate overlap configuration provides a highly robust, jitter-free geometric transition without unnecessarily sacrificing inference efficiency.

\begin{wraptable}{r}{0.28\textwidth}
 \vspace{-1.6em}
 \centering
    \caption{Analysis of LoRA ranks.}
    \vspace{-1em}
    \renewcommand\tabcolsep{7.2pt}
    \renewcommand\arraystretch{1.2}
    \footnotesize
    \label{app_tab:ablation_lora}
    \begin{tabular}{c|cc}
        \hlineB{2.5}
        \rowcolor{CadetBlue!20} 
        \textbf{LoRA Rank} & \textbf{AbsRel$\downarrow$} & \textbf{$\delta_1\uparrow$}  \\
        \hlineB{1.5}
        256   & 7.7 & 0.974  \\
        \rowcolor{gray!10}
        512 & 7.3 & 0.977 \\
        1024 & 7.3 & 0.979   \\
        \hlineB{2.5}
    \end{tabular}
    \vspace{-1.4em}
\end{wraptable}

\noindent\textbf{Analysis of LoRA Rank.}\quad
Unlike standard LoRA applications that employ low ranks for superficial style transfer, adapting a video diffusion backbone into a dense geometric regressor requires modeling highly complex mappings. Table~\ref{app_tab:ablation_lora} ablates the LoRA capacity on ScanNet. We observe that a moderate rank of $256$ yields sub-optimal accuracy, while expanding to rank $512$ significantly improves structural fidelity (AbsRel drops to $7.3$). Further scaling to rank $1024$ provides only marginal gains. Empirically, we found that extremely low ranks struggle to capture high-frequency details, whereas full parameter fine-tuning tends to overfit the limited training data and degrade the model's pre-trained zero-shot priors. Consequently, rank $512$ offers an optimal balance between geometric capacity and prior preservation.

\begin{figure}[!t]
    \centering
    \includegraphics[width=\linewidth]{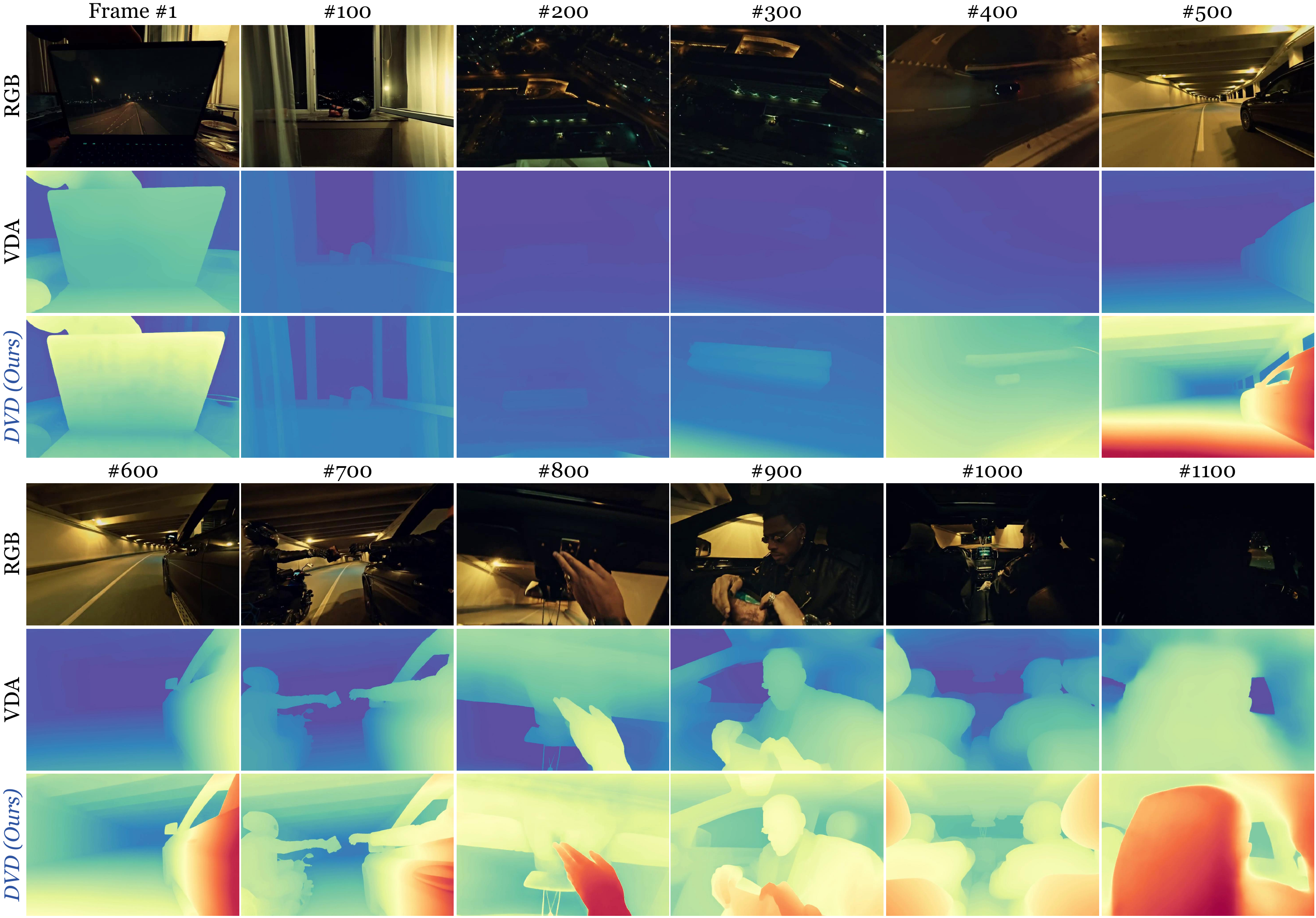}
    \vspace{-2em}
    \caption{
     Failure case analysis on $1100$-frame sequence with massive scene transitions. While both \ourmethod and VDA~\citep{chen2025video} inevitably suffer from global scale drift across disjointed scenes (\textit{e.g.}, contrasting the absolute depth representations between Frame \#$1$ and \#$500$), \ourmethod consistently maintains significantly sharper local structural fidelity (\textit{e.g.}, hands in Frames \#$800$--\#$900$).
    }
    \label{app_fig:failure_case}
    \vspace{-0.8em}
\end{figure}

\section{Failure Case Analysis}
\label{app:failure}
As discussed in Section~\S\ref{apx:limitations}, the empirical affine coherence of \ourmethod relies on geometric overlap between adjacent temporal windows. Consequently, the model may experience scale inconsistencies in unconstrained long-video scenarios characterized by massive ego-motion or abrupt scene transitions.

Figure~\ref{app_fig:failure_case} illustrates a highly challenging $1100$-frame sequence featuring drastic environmental shifts (\textit{e.g.}, transitioning from an indoor desk to an outdoor tunnel). When comparing distant frames with zero visual overlap (\textit{e.g.}, Frame \#$1$ \textit{vs.} Frame \#$500$), \ourmethod inevitably exhibits global scale drift, struggling to anchor a unified absolute depth range across completely disjointed scenes. 

Crucially, however, this limitation is an open challenge for the entire field rather than a specific flaw of our paradigm. The state-of-the-art discriminative baseline, VDA~\citep{chen2025video}, suffers from identical, if not more severe, temporal scale degradation under these extreme dynamics. Furthermore, even amidst global scale drift, \ourmethod consistently preserves vastly superior high-frequency structural fidelity (\textit{e.g.}, the sharp laptop screen in Frame \#$1$ and the intricate hand geometries in Frames \#$800$--\#$900$) compared to the over-smoothed predictions of VDA. This confirms that while infinite-length metric anchoring remains an unresolved problem, our deterministic generative prior still guarantees unparalleled local geometric precision.

\section{Exhibition Board}
\label{apx:viz}

To further demonstrate the robust zero-shot generalization of \ourmethod across highly diverse open-world domains, this section presents an extensive exhibition of qualitative results. Our visualizations encompass a wide spectrum of scenarios, including natural landscapes, complex architecture, dynamic subjects (humans and animals), as well as out-of-domain stylized content such as animations, video games, and AI-generated videos. Detailed predictions for short video clips are provided in Figures~\ref{app_fig:demonstration_short_1},~\ref{app_fig:demonstration_short_2}, and~\ref{app_fig:demonstration_short_3}, while unconstrained long-video results are showcased in Figures~\ref{app_fig:demonstration_long_1},~\ref{app_fig:demonstration_long_2}, and~\ref{app_fig:demonstration_long_3}.

\begin{figure}[!t]
    \centering
    \vspace{-2.4em}
    \includegraphics[width=0.86\linewidth]{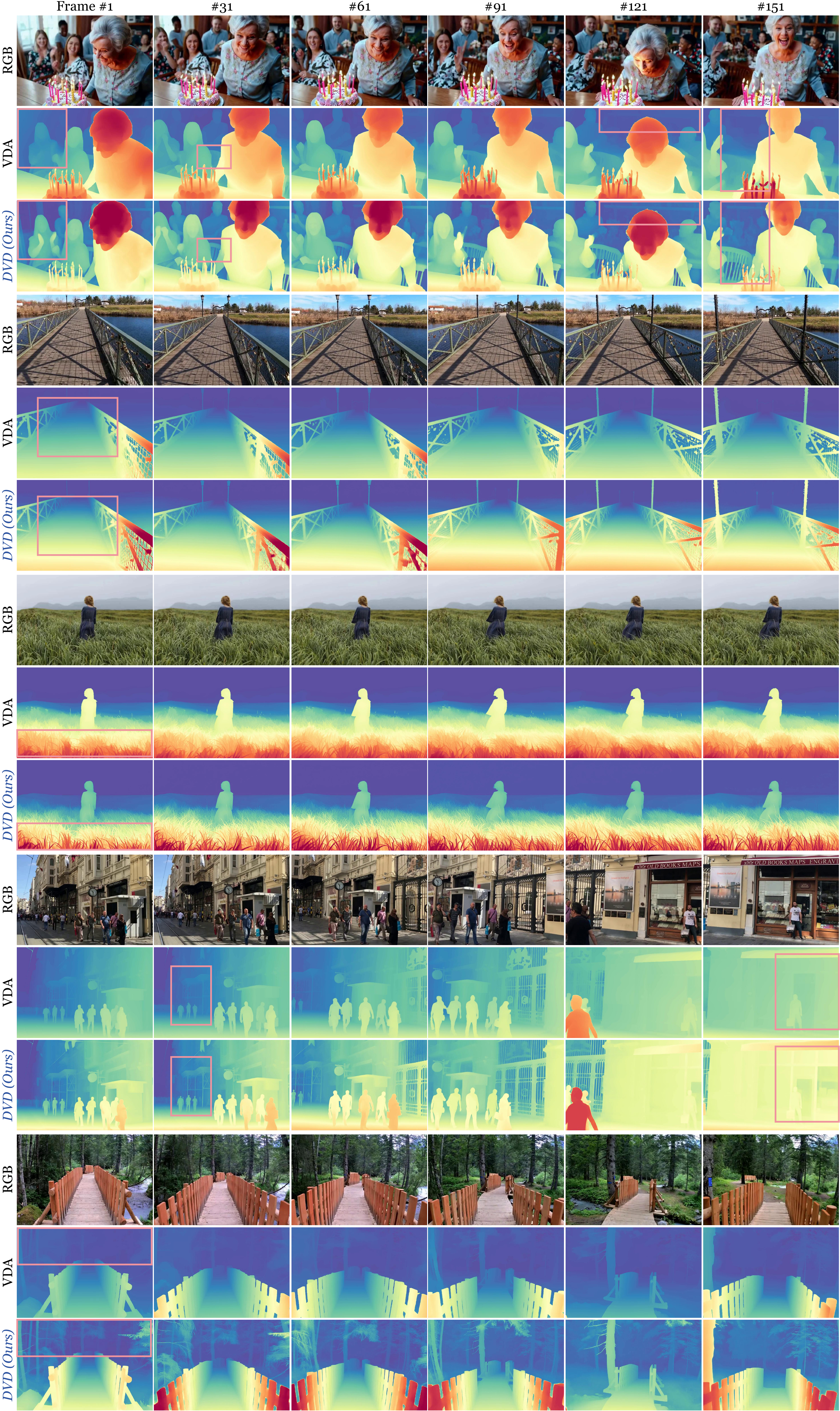}
    \vspace{-1em}
    \caption{
     More results demonstrations on short videos.
    }
    \label{app_fig:demonstration_short_1}
    \vspace{-1.6em}
\end{figure}

\begin{figure}[!t]
    \centering
    \vspace{-2.4em}
    \includegraphics[width=0.86\linewidth]{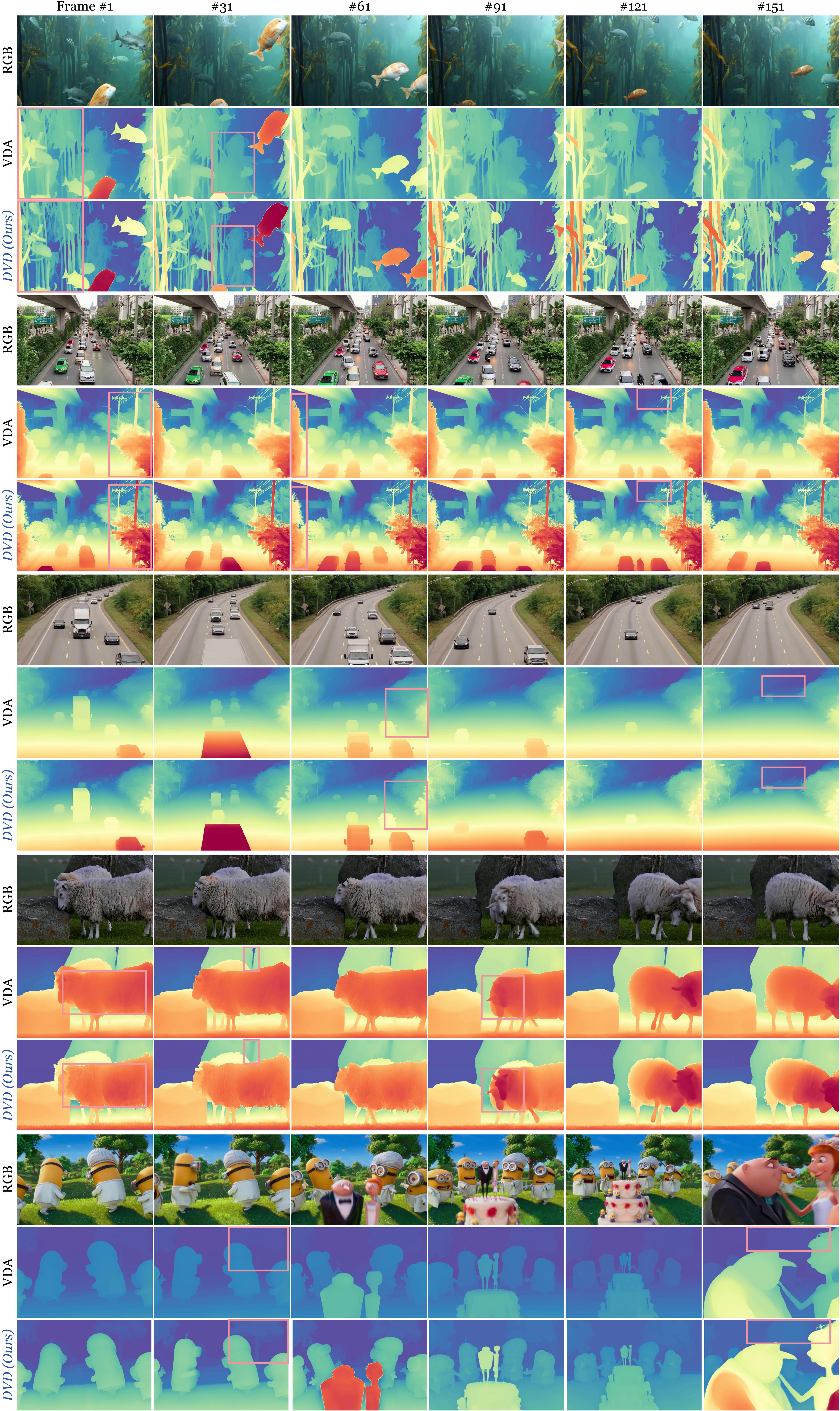}
    \vspace{-1em}
    \caption{
     More results demonstrations on short videos.
    }
    \label{app_fig:demonstration_short_2}
    \vspace{-1.6em}
\end{figure}

\begin{figure}[!t]
    \centering
    \vspace{-2.4em}
    \includegraphics[width=0.86\linewidth]{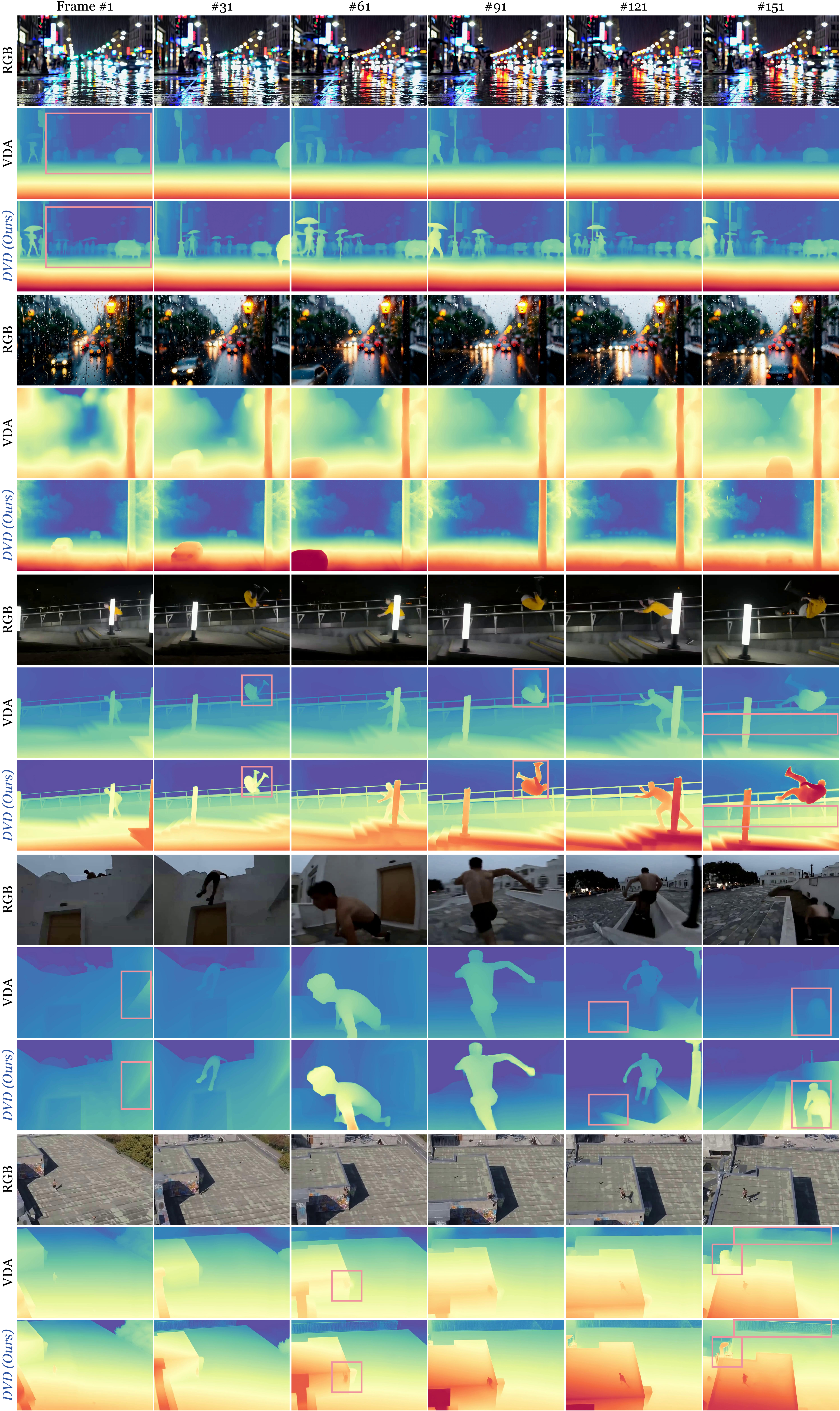}
    \vspace{-1em}
    \caption{
     More results demonstrations on short videos.
    }
    \label{app_fig:demonstration_short_3}
    \vspace{-1.6em}
\end{figure}

\begin{figure}[!t]
    \centering
    \includegraphics[width=\linewidth]{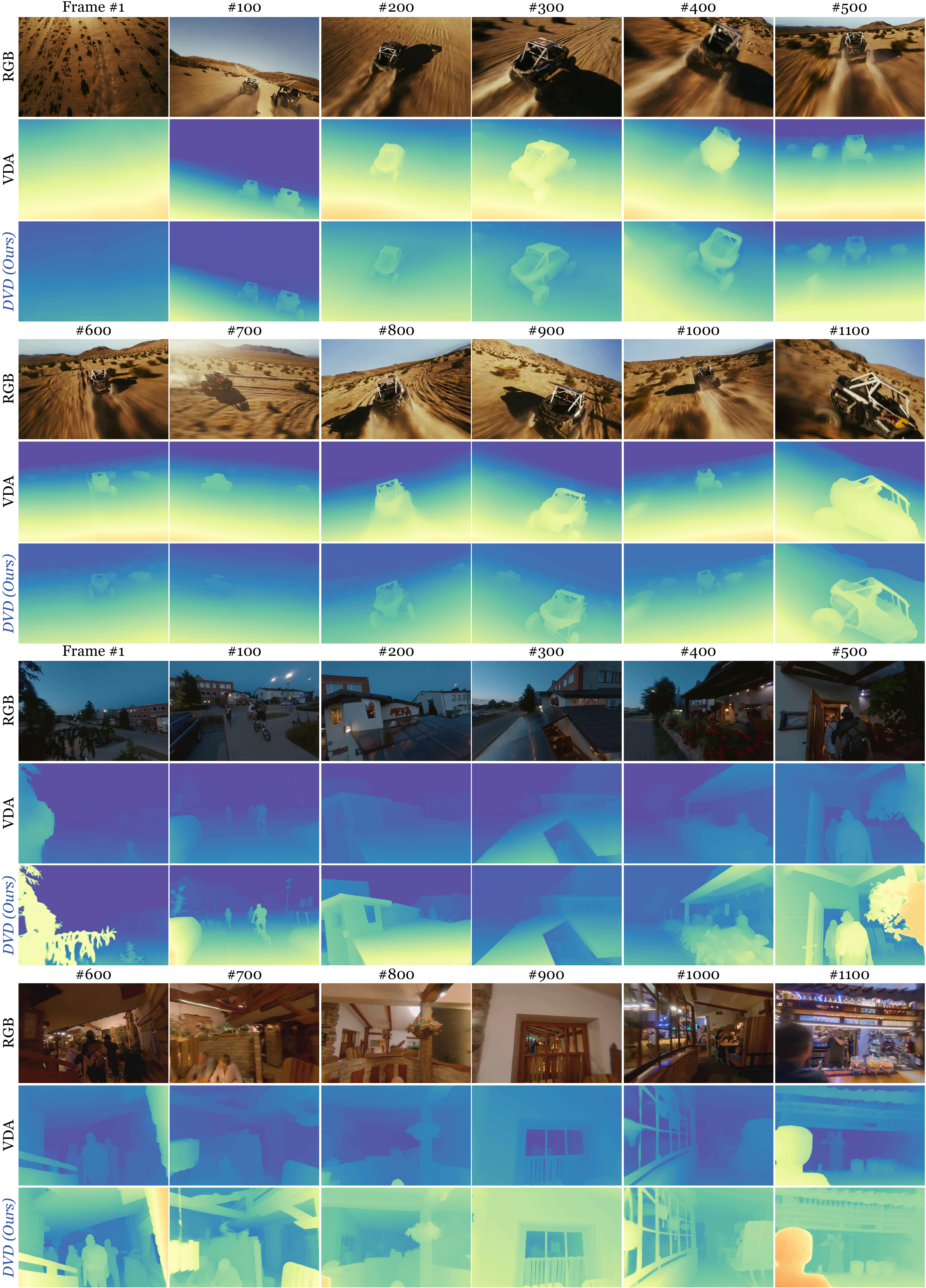}
    \vspace{-2em}
    \caption{
     More results demonstrations on long videos.
    }
    \label{app_fig:demonstration_long_1}
    \vspace{-1.6em}
\end{figure}

\begin{figure}[!t]
    \centering
    \includegraphics[width=\linewidth]{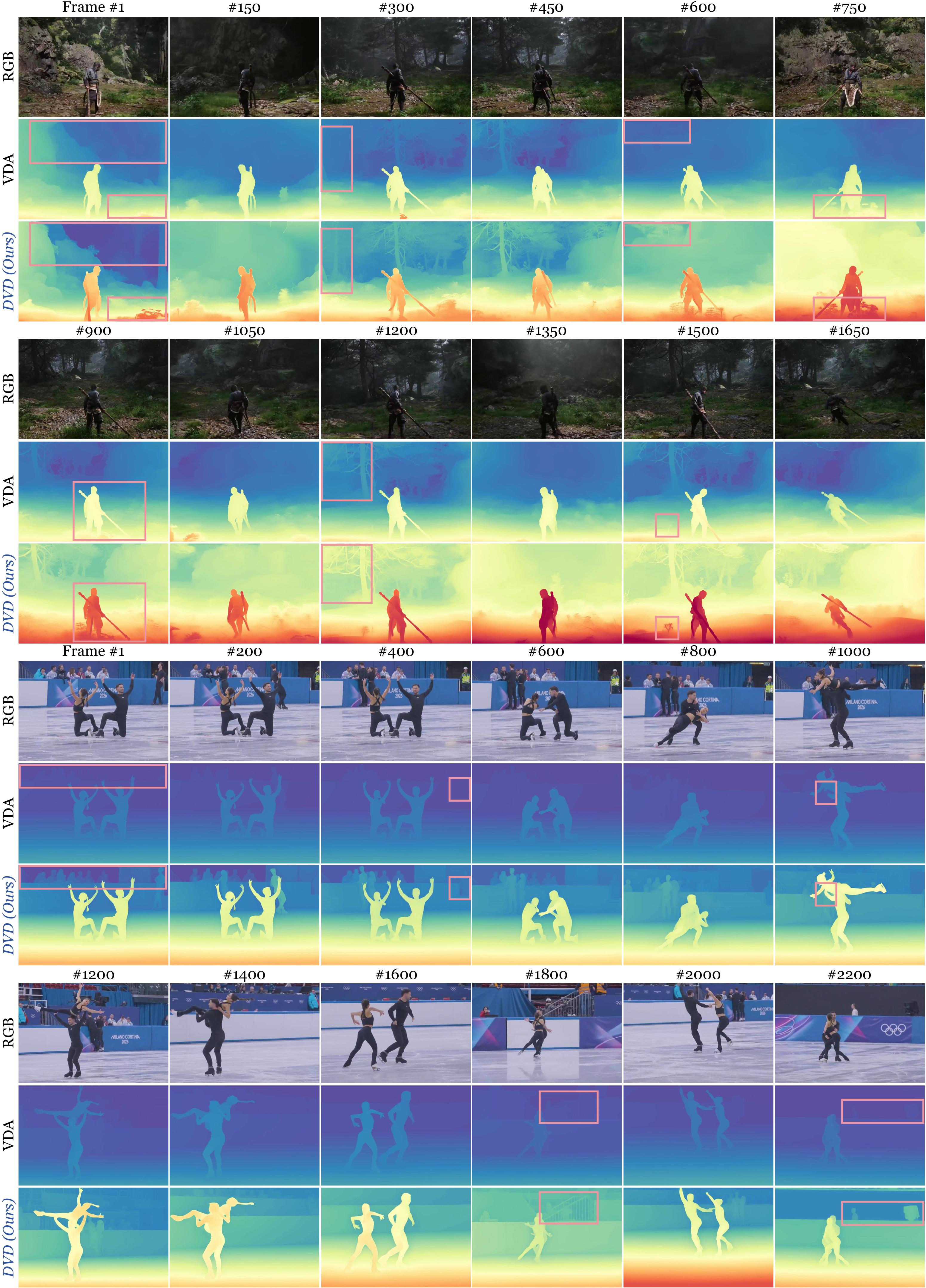}
    \vspace{-2em}
    \caption{
     More results demonstrations on long videos.
    }
    \label{app_fig:demonstration_long_2}
    \vspace{-1.6em}
\end{figure}

\begin{figure}[!t]
    \centering
    \includegraphics[width=\linewidth]{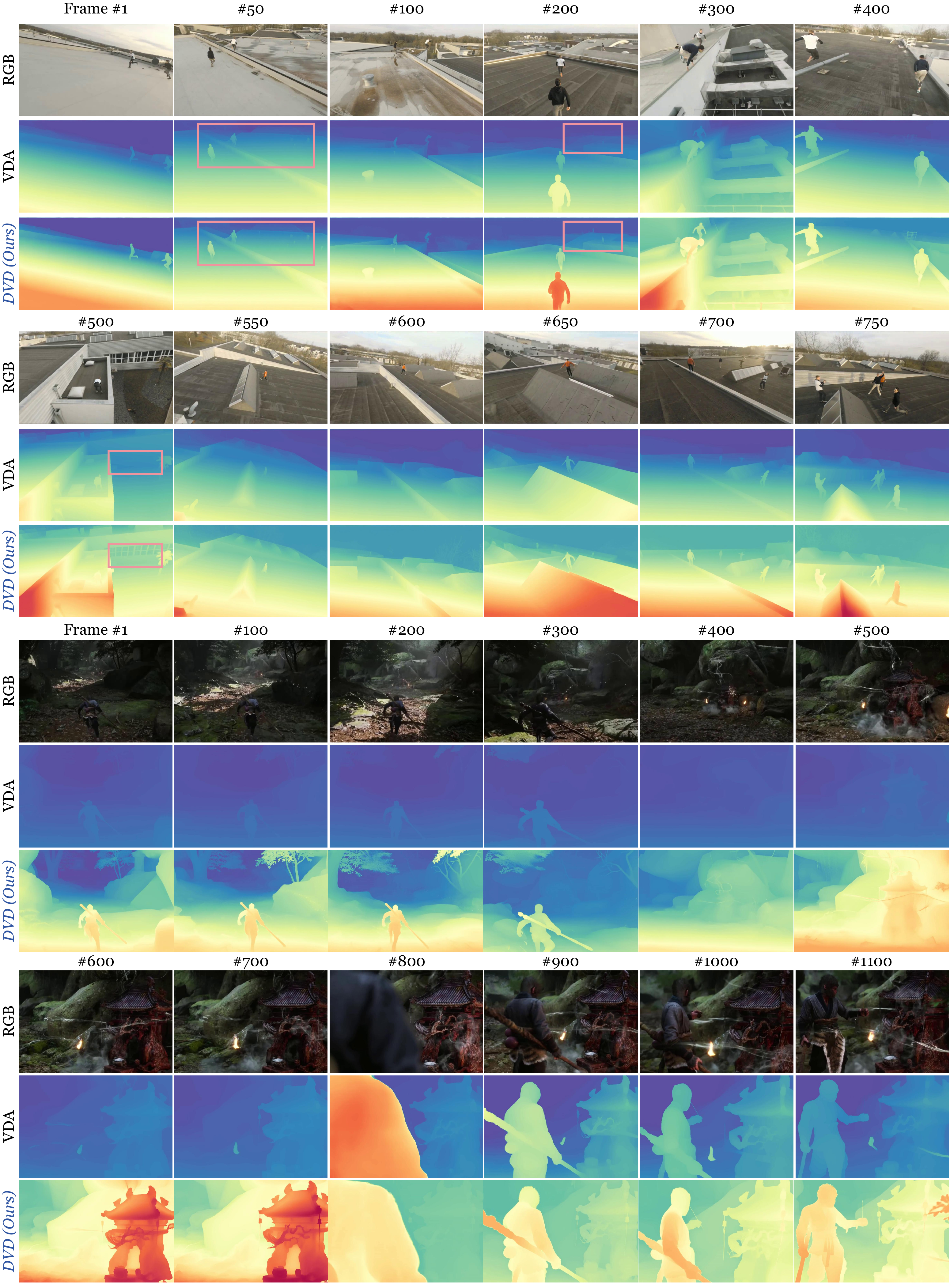}
    \vspace{-2em}
    \caption{
     More results demonstrations on long videos.
    }
    \label{app_fig:demonstration_long_3}
    \vspace{-1.6em}
\end{figure}


\end{document}